\documentclass[11pt,letterpaper]{article}
\usepackage[margin=1in]{geometry}
\usepackage[T1]{fontenc}
\usepackage{lmodern,graphicx,booktabs,array,caption,amsmath,enumitem,xcolor}
\usepackage[numbers,sort&compress]{natbib}
\usepackage[colorlinks,linkcolor=blue!45!black,citecolor=blue!45!black,urlcolor=blue!45!black]{hyperref}
\definecolor{familyband}{HTML}{DDE7FF}
\newcommand{\familyheading}[2]{%
  \noalign{\begingroup\color{familyband}\hrule height\dimexpr\ht\strutbox+\dp\strutbox\relax\endgroup
    \vskip\dimexpr-\ht\strutbox-\dp\strutbox\relax}%
  \multicolumn{#1}{l}{\strut\textbf{\textit{#2}}}}
\definecolor{answerone}{HTML}{EAF0FA}
\definecolor{answertwo}{HTML}{E8F3ED}
\definecolor{casefinding}{HTML}{FBF0DA}
\newcommand{\casepanel}[3]{\par\noindent\begingroup\setlength{\fboxsep}{5pt}%
  \colorbox{#1}{\parbox{\dimexpr\linewidth-2\fboxsep\relax}{\small\raggedright
    \textbf{#2}\par #3}}\endgroup\par\smallskip}
\definecolor{recallgain}{HTML}{E3F0E6}
\definecolor{recallloss}{HTML}{F8E5E5}
\newcommand{\recalldelta}[1]{\begingroup
  \ifdim #1pt>0pt\def\shade{recallgain}\else
  \ifdim #1pt<0pt\def\shade{recallloss}\else\def\shade{white}\fi\fi
  \setlength{\fboxsep}{0pt}\colorbox{\shade}{\makebox[3.9em][r]{\strut #1\hspace{.25em}}}\endgroup}
\title{WildAIGC-Bench: Can AI Image Detectors Handle the Wild?}
\author{Zijian Yu\textsuperscript{1} \and Hangwei Zhang\textsuperscript{1} \and Shunxing Chen\textsuperscript{1} \and Jun Lan\textsuperscript{1} \and Huijia Zhu\textsuperscript{1} \and Weiqiang Wang\textsuperscript{1} \and Jianfu Zhang\textsuperscript{2}}
\date{}
\begin{document}
\maketitle
\begin{center}\small
\textsuperscript{1}Ant Group \qquad \textsuperscript{2}Shanghai Jiao Tong University\par
Corresponding author: Jun Lan (\href{mailto:yelan.lj@antgroup.com}{yelan.lj@antgroup.com})\par
Contact: Zijian Yu (\href{mailto:yuzijian.yzj@antgroup.com}{yuzijian.yzj@antgroup.com})
\end{center}
\begin{abstract}
An AI-generated scene or a subtle edit to a photograph can make a false claim about a public event appear credible. Existing benchmarks test detectors across generators and platforms, but offer limited insight into how well they recognize fabrications that circulate as evidence for such claims. We introduce WildAIGC-Bench, an event-driven benchmark designed for broad coverage, timely updates, and human-verified labels. A recurring search across news outlets, fact-checking sites, and social posts identifies emerging cases, and reviewers verify the evidence for each image. The benchmark contains 600 AI-generated or edited images and 638 genuine photographs of related subjects. It preserves posts alongside crops of the depicted images and links documented edits to their original photographs where available. All ten visual detectors achieve lower balanced accuracy on WildAIGC-Bench than their respective averages across eight external benchmarks with both classes. On 37\% of the AI images, at least three quarters of the evaluated detectors predict real. Cropping improves recall for most visual detectors but reduces it for most general-purpose multimodal models. Comparisons with editing originals further reveal local alterations that most detectors miss even when they correctly accept the original photograph.
\end{abstract}
\begin{center}\small Dataset: \url{https://github.com/YuZijian/WildAIGC-Bench}\end{center}
\section{Introduction}
Photographs shared online often serve as evidence for claims about public events. AI generation can give a fictitious event the appearance of a press photograph, while editing a small part of a genuine image can change who appears to have been present or what appears to have happened. Both forms of fabrication can retain the visual detail that makes a photograph credible. Assessing detectors in this setting therefore requires images that have actually circulated in connection with such claims.
\begin{figure}[t]\centering\includegraphics[width=.96\textwidth]{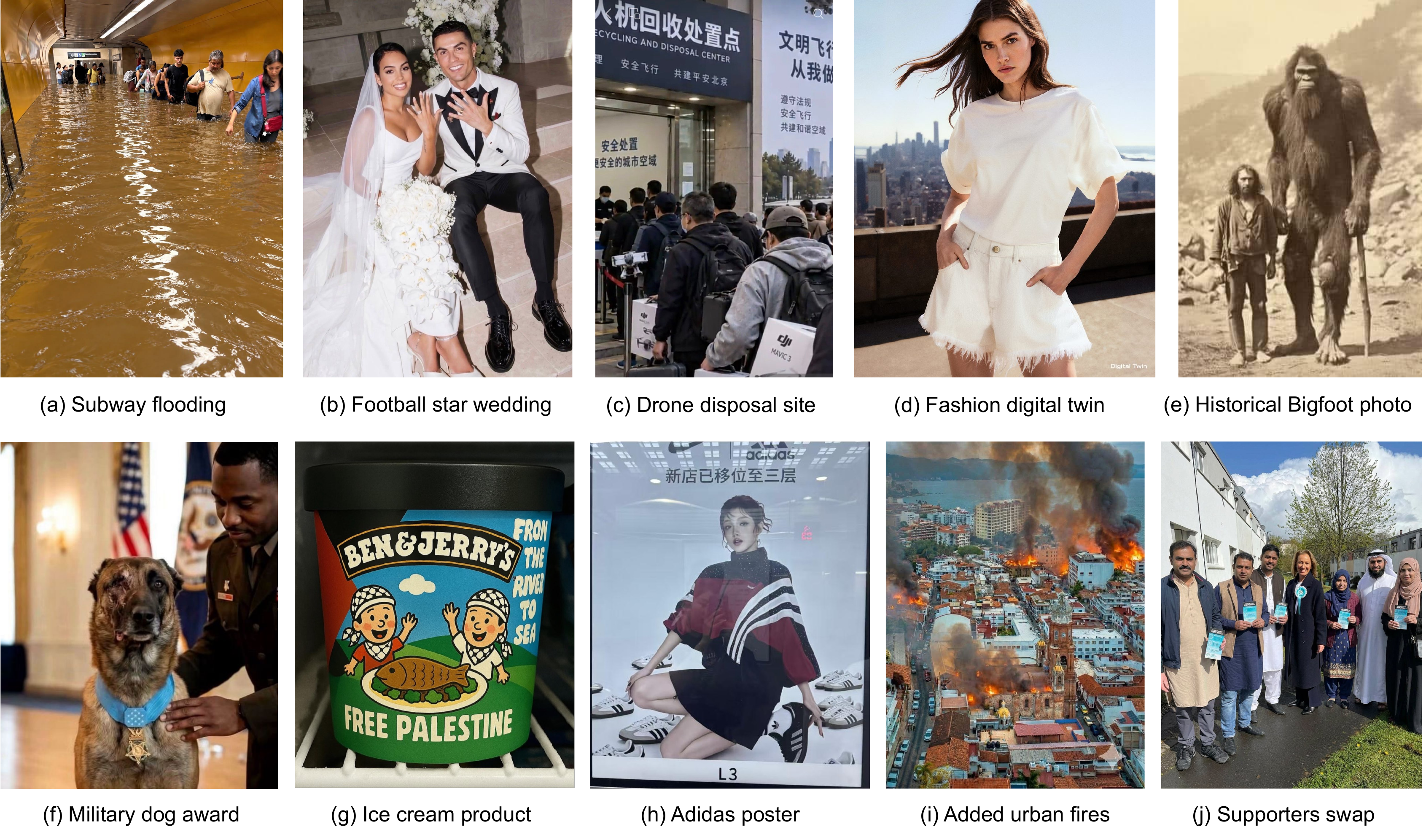}
\caption{Images encountered in public circulation. The collection includes fabricated events, plausible product and fashion imagery, and changes to specific parts of an existing photograph. Panels (i) and (j) illustrate how a limited change can alter what a photograph appears to document. Sources and case descriptions appear in Appendix B.}\label{fig:examples}
\end{figure}

Existing benchmarks have made it possible to measure generalization across generators~\cite{univfd,genimage}, assess realistic images from creative communities~\cite{aide_chameleon}, and evaluate images acquired from social platforms~\cite{clip_synthwildx,wildrf,itw_sm}. Where the images come from shapes what these tests can reveal. Collecting from a generator or an account that publishes synthetic work gives a direct route to positive labels. A small edit to a photograph may instead come to light days after publication, when a reporter locates the original or the creator explains how it was altered. Both kinds of image occur online, but the latter can be difficult to discover by searching for examples of a particular generator. Whether high recall on existing collections transfers to these cases is an empirical question.

We introduce WildAIGC-Bench, a benchmark built by following the events and claims through which images circulate. Figure~\ref{fig:examples} illustrates the range of images collected, from fabricated scenes to local edits of existing photographs. Scheduled tasks identify candidate images across news outlets, fact-checking sites, and related social posts, then screen them and prepare preliminary annotations. The distinction between a false claim and a fabricated image is central to this process. A genuine photograph shared with the wrong date can support a false story without containing any AI-generated content. Reviewers therefore match the evidence to the target image before admitting a positive example. A creator's disclosure, a documented comparison with an editing original, or a published investigation can establish AI involvement even when the generating model is unknown.

The collection targets a specific use case: assessing an image that has circulated as photographic evidence for a claim. Inclusion depends on finding a source that documents how the image was made or altered. The resulting sample therefore emphasizes traceable cases, and its class balance is set by collection. The photographs serve three purposes. Photographs of similar subjects measure false alarms on the kinds of scenes a detector encounters here. A photograph of the same event can document an unusual scene from another viewpoint. An editing original gives a stronger control: the detector must distinguish two versions of the same scene. For images embedded in posts, we also retain the acquired file and the target crop to measure how decisions change when surrounding material is removed.

We evaluate visual detectors, specialist multimodal detectors, and general-purpose multimodal models. All ten visual configurations have lower balanced accuracy here than their respective means on eight external benchmarks containing both AI images and photographs. Moving to a different detector family recovers some of those misses, but substantial overlap remains. The explanations also reveal why similar-looking decisions deserve closer inspection: a model may detect an edit through a visual inconsistency, recognize a previously circulated fabrication, or invoke the history of an event to assess whether the depicted scene is plausible. Our contributions are:
\begin{itemize}[topsep=2pt,partopsep=0pt,itemsep=1pt,parsep=0pt]
\item \textbf{A continuous, event-driven construction pipeline.} Scheduled retrieval, screening and preliminary annotation turn online cases into evidence-linked records for human verification and quality control, supporting repeatable dataset expansion.
\item \textbf{A benchmark with broad coverage, timely updates, and human-verified labels.} WildAIGC-Bench contains 600 circulating AI-generated or edited images and 638 genuine photographs, with source evidence, editing originals and paired image views.
\item \textbf{Comprehensive evaluation and failure analysis.} Evaluating 24 configurations reveals substantial transfer gaps, shared blind spots, and the influence of local edits, context and event knowledge.
\end{itemize}

\section{Related Work}
\label{sec:related}
\subsection{AIGC Detection}
Visual detectors use several complementary approaches to identify AI-generated images. Artifact-based learning captures transferable generation traces through augmentation or reconstruction-based contrastive training~\cite{cnndetection,drct}. Representation-based approaches use pretrained features or combine semantic and low-level cues for classification~\cite{univfd,aide_chameleon,effort,omniaid,cospy}. PGC uses the strongest local cues to calibrate the image-level decision~\cite{pgc}. Detector generalization has also been studied through blur robustness~\cite{dblur}, transferable features~\cite{globalforge,lowrank}, and adaptation to new domains~\cite{fleet,hydra}. Data-centric approaches align the content and processing characteristics of real and synthetic training images, reducing shortcuts unrelated to generation~\cite{bfree,dda_evalgen}.

MLLM-based detection combines authenticity prediction with language-based analysis. Explanation-supervised approaches learn from artifact descriptions to connect predictions with visual evidence~\cite{fakevlm}. UniGenDet jointly optimizes image generation and detection~\cite{unigendet}. Reasoning-oriented approaches use structured supervision or reinforcement learning to develop evidence assessment, including planning, self-reflection, and reasoning across image and video inputs~\cite{veritas,busterxpp}. Perception-oriented approaches strengthen sensitivity to fine details, semantic anomalies, and pixel-level differences, then integrate these abilities into reasoning through distillation~\cite{veritaspp}. DeepVRM injects low-level forensic features into an intermediate model layer~\cite{deepvrm}. WildAIGC-Bench tests whether these capabilities transfer to generated scenes and edits documented in public circulation.

\subsection{AIGC Detection Benchmarks}
CNNDetection~\cite{cnndetection} and UnivFD~\cite{univfd} support evaluation across generation architectures. GenImage provides large-scale cross-generator and degradation tests~\cite{genimage}; Synthbuster focuses on diffusion imagery~\cite{synthbuster}; and EvalGEN includes newer generation architectures~\cite{dda_evalgen}. WildFake supports training and generalization analysis across diverse content and hierarchically organized generator types, architectures, weights, and versions~\cite{wildfake}. AIGIBench broadens task coverage and includes social-network and creative-community content~\cite{aigibench}. Many standard evaluations rely on clean generator outputs. Their results may not reflect detection after compression, screenshots, and recapture in real use; predefined degradation tests cover only selected aspects of these conditions.

\subsection{In-the-Wild Benchmarks}
\emph{In-the-wild} benchmarks target real-use conditions through online image collection or realistic sharing processes. Chameleon emphasizes realistic content from ArtStation, Civitai, and Liblib~\cite{aide_chameleon}. SynthWildX collects images from X~\cite{clip_synthwildx}, while WildRF uses social-network imagery~\cite{wildrf}. ITW-SM covers multiple social platforms, obtaining AI images from accounts and communities that openly share generated content~\cite{itw_sm}. MIRAGE includes online news, social media, and e-commerce scenarios alongside pipelines that combine multiple generation steps~\cite{mirage}. RRDataset examines actual platform transmission, screen capture, and recapture~\cite{rrdataset}. NTIRE-RobustAIGenDetection~\cite{ntire2026wild} includes real and AI-generated images with transformations such as resizing, blurring, and compression to simulate online sharing.

Creator and AI-community collections can favor works showcasing generation capabilities, whose content may differ from everyday photographs despite appearing realistic. Controlled sharing and recapture tests cover selected transmission conditions. Real circulating forgeries combine photographic realism, local manipulation and heterogeneous dissemination histories, creating a harder detection distribution that existing in-the-wild collections do not fully represent. WildAIGC-Bench instead follows actual news events and misleading claims to collect photographic forgeries, preserving their acquired condition. For AI images, the source link and evidence summary connect the label to a particular forgery. Credited photographs provide controls for the associated false-positive risk.

\begin{figure*}[t]
\centering\includegraphics[width=.95\textwidth]{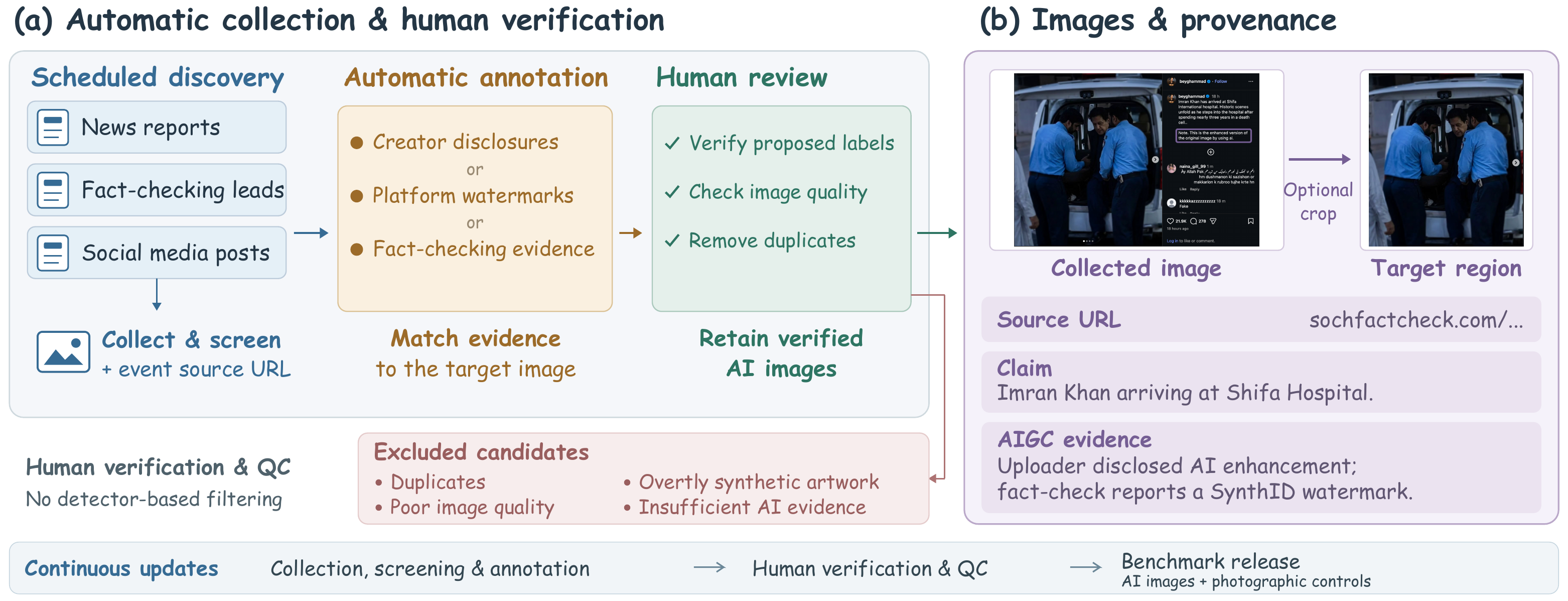}
\caption{WildAIGC-Bench construction. Scheduled tasks identify candidates and collect image-specific evidence for human review. The example record links a circulating post to the cropped image and the reported evidence of AI enhancement.}\label{fig:workflow}
\end{figure*}
\subsection{Local Edits and Contextual Evidence}
A local edit can change the meaning of a photograph while leaving most pixels intact. Detectors address this problem by examining patches~\cite{patch,panoptic} or preserving fine detail during resizing~\cite{nopixel,spectral}. Localization methods identify altered regions~\cite{zoom,regflow,shield,sida}. Our edit--original pairs test whether the same detector flags an edit and accepts its source photograph.

Explanatory detectors connect verdicts to image regions~\cite{legion,vigil}, describe visual artifacts~\cite{holmes,see,reveal}, or reason about semantic inconsistencies~\cite{semantic,tranx,dfllava}. Other approaches retrieve and combine evidence~\cite{defake,fox}. Our paired posts and crops let us examine a different source of evidence: a watermark or caption can help a model recognize a fabrication even when the depicted scene appears plausible.

\section{WildAIGC-Bench Construction}
WildAIGC-Bench is built through a continuously operating workflow for automated retrieval, screening and preliminary annotation, followed by human verification and quality control (Fig.~\ref{fig:workflow}). The workflow delivers broad coverage across online sources, timely updates as new cases emerge, and human-verified labels grounded in image-specific evidence. Its event-driven design captures both generated scenes and subtle edits circulating as photographic evidence.

\subsection{Collection Pipeline}
Collection follows public events and online claims across news websites, fact-checking outlets and associated social posts. These sources provide complementary routes to discovery: a post exposes the circulating image, a report establishes its context, and a creator disclosure or investigation supplies evidence of generation or editing. Following these connections expands coverage across subjects, platforms and manipulation types, including cases whose generating models are unknown. The collection includes both images circulated to support misleading event claims and openly disclosed AI imagery, such as generated photographs used in commercial campaigns.

Scheduled searches use GPT, Gemini, and Qwen to find candidate images. GPT screens the candidates for relevance and prepares a preliminary annotation for each retained image. Each candidate record associates the target image with its source page, the claim it accompanies and the available evidence of AI involvement. The annotation summarizes creator disclosures, comparisons with original photographs, visible generation marks and published verification findings. Organizing this information around individual images gives reviewers a concrete evidence record to inspect and supports consistent processing across heterogeneous websites.

This recurring workflow keeps discovery active as new events, generators and editing practices appear. Automation handles retrieval and provisional annotation, allowing human effort to focus on label verification and final admission. Because discovery follows documented cases, it can incorporate unfamiliar generation techniques without requiring advance knowledge of a generator or its characteristic artifacts. No AI-image detector is used to select samples. The same workflow supports continued expansion of the benchmark and the addition of related authentic photographs.

\subsection{Verification and Quality Control}
Human reviewers verify every provisional label against evidence tied to the target image. They inspect the cited disclosure, original photograph, generation mark or investigation and confirm that it establishes AI generation or editing for that particular image. This image-level check distinguishes fabricated visual content from authentic photographs accompanied by false claims. It also separates disputed images from genuine references reproduced on the same webpage, preserving the correct label for each target.

Evidence remains traceable through source links and supporting summaries. Invisible-watermark evidence is attributed to the fact-checking organization that published the detection result. In the hospital-arrival example in Fig.~\ref{fig:workflow}, the acquired post contains an AI-enhancement disclosure, while the linked investigation reports a watermark finding. These complementary records support verification even when the target image itself looks photographically plausible.

Reviewers then perform quality control and decide final inclusion. Duplicate candidates, insufficiently supported labels, unusable files and obviously stylized artwork are excluded. Screenshots, compressed images and other circulation artifacts are retained when the target remains legible. This preserves the conditions under which people actually encounter the images while grounding admission in verifiable evidence. Human review turns automatically discovered candidates into a curated evaluation set with auditable labels.

\subsection{Dataset Composition and Paired Views}
\begin{figure}[t]
\centering
\includegraphics[width=.66\linewidth]{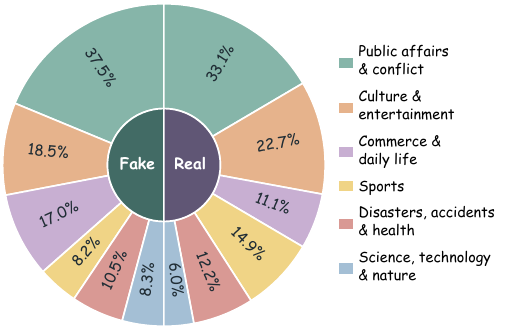}
\caption{Topic distribution of 600 AI images and 638 real photographs. Each semicircle represents 100\% of its class, with topics in the same order from top to bottom.}\label{fig:topics}
\end{figure}

The benchmark contains 600 AI-generated or edited images from 487 primary source pages across 162 hostnames, alongside 638 real photographs from 89 pages across 34 hostnames. Real images are selected from themes represented among the fabrications and include both ordinary scenes and unusual public events. Photographer credits, editorial records and archival attribution support their provenance. Figure~\ref{fig:topics} compares the two classes by the primary subject of each image, assigned from its recorded title and source context. Public affairs and conflict form the largest category in both classes. Images of commerce and everyday life account for 17.00\% of AI images and 11.13\% of real photographs. The pattern reverses for sports (8.17\% and 14.89\%, respectively). The controls thus cover the same broad subjects with different frequencies. Appendix B details source coverage.

The dataset also preserves documented relationships between AI images and authentic photographs. Twenty edits are linked to 18 distinct originals. Figure~\ref{fig:editing-original} shows a concrete example: removing parked vehicles changes the apparent evidence of a fire-safety violation while preserving most of the scene. These pairs test sensitivity to manipulation within a common photographic setting. Another 24 links connect 20 AI images to 21 genuine photographs of the same event, providing contextual references for evaluating fabricated depictions of real events.

Each sample retains the acquired online file (raw). When a post or collage contains surrounding text or interface elements, a target-region crop provides the processed view. There are 126 AI images with distinct raw and processed views; the remaining 474 AI images and all 638 real photographs use identical files across views. These paired inputs reveal how target scale, surrounding text and platform cues affect detection. Together, authentic controls, editing originals and paired views support analyses that extend beyond aggregate detection accuracy.
\begin{table}[t]\centering\small
\caption{Detection on WildAIGC-Bench (percent). FPR is shared across views; BAcc uses processed inputs. The right block reports recall on changed AI pairs with valid decisions in both views. $\Delta$ is processed minus raw recall in points. Within each family, bold marks the two highest processed recalls and BAcc scores and the two lowest FPRs, including ties.}\label{tab:detectors}
\setlength{\tabcolsep}{2.4pt}
\begin{tabular*}{\textwidth}{@{\extracolsep{\fill}}lrrrrrrr@{}}\toprule
 & \multicolumn{4}{c}{Full benchmark (1,238 images)} & \multicolumn{3}{c}{Changed AI pairs (126)}\\
\cmidrule(lr){2-5}\cmidrule(l){6-8}
Model & Raw recall & Proc. recall & FPR $\downarrow$ & BAcc $\uparrow$ & Raw recall & Proc. recall & $\Delta$\\\midrule
\familyheading{8}{Visual detectors}\\
DRCT~{\footnotesize [ICML 2024]} & 29.83 & 29.67 & 14.11 & 57.78 & 32.54 & 31.75 & \recalldelta{-0.79}\\
AIDE~{\footnotesize [ICLR 2025]} & 16.50 & 21.33 & 12.38 & 54.48 & 11.90 & 34.92 & \recalldelta{+23.02}\\
B-Free~{\footnotesize [CVPR 2025]} & 35.83 & \textbf{36.33} & \textbf{3.13} & \textbf{66.60} & 25.40 & 27.78 & \recalldelta{+2.38}\\
CO-SPY~{\footnotesize [CVPR 2025]} & 31.50 & 33.50 & 26.49 & 53.51 & 27.78 & 37.30 & \recalldelta{+9.52}\\
Effort~{\footnotesize [ICML 2025]} & 27.33 & 28.00 & 20.69 & 53.66 & 29.37 & 32.54 & \recalldelta{+3.17}\\
DDA~{\footnotesize [NeurIPS 2025]} & 37.33 & \textbf{37.67} & 8.46 & \textbf{64.60} & 32.54 & 34.13 & \recalldelta{+1.59}\\
OmniAID~{\footnotesize [ICML 2026]} & 30.00 & 30.83 & 5.80 & 62.52 & 25.40 & 29.37 & \recalldelta{+3.97}\\
PGC~{\footnotesize [ICML 2026]} & 15.67 & 15.50 & 14.58 & 50.46 & 17.46 & 16.67 & \recalldelta{-0.79}\\
PE-336 & 19.67 & 20.50 & 8.62 & 55.94 & 9.52 & 13.49 & \recalldelta{+3.97}\\
PE-518 & 22.00 & 22.83 & \textbf{1.72} & 60.55 & 10.32 & 14.29 & \recalldelta{+3.97}\\
\midrule
\familyheading{8}{Forensic specialist MLLMs}\\
FakeShield~{\footnotesize [ICLR 2025]} & 31.00 & 30.33 & 11.76 & 59.29 & 37.30 & 34.13 & \recalldelta{-3.17}\\
FakeVLM~{\footnotesize [NeurIPS 2025]} & 64.50 & \textbf{66.50} & 49.53 & 58.49 & 55.56 & 65.08 & \recalldelta{+9.52}\\
UniGenDet~{\footnotesize [CVPR 2026]} & 43.00 & \textbf{45.00} & \textbf{9.56} & \textbf{67.72} & 30.95 & 40.48 & \recalldelta{+9.52}\\
BusterX++~{\footnotesize [2026 release version]} & 23.83 & 24.67 & 14.11 & 55.28 & 20.63 & 24.60 & \recalldelta{+3.97}\\
DeepVRM~{\footnotesize [ICML 2026]} & 30.17 & 30.17 & 18.65 & 55.76 & 30.16 & 30.16 & \recalldelta{0.00}\\
Veritas++~{\footnotesize [arXiv 2026]} & 33.67 & 35.33 & \textbf{3.13} & \textbf{66.10} & 15.87 & 23.81 & \recalldelta{+7.94}\\
\midrule
\familyheading{8}{General-purpose MLLMs}\\
Qwen3.8-27B & 55.50 & 54.50 & 12.38 & 71.06 & 54.76 & 50.00 & \recalldelta{-4.76}\\
Qwen3.6-35B-A3B & 39.17 & 37.17 & 1.88 & 67.64 & 52.38 & 42.86 & \recalldelta{-9.52}\\
GLM-5.3-Flash & 50.00 & 46.50 & \textbf{1.41} & 72.54 & 61.11 & 44.44 & \recalldelta{-16.67}\\
MiniMax-M3 & 61.50 & \textbf{61.50} & 16.61 & 72.44 & 62.70 & 62.70 & \recalldelta{0.00}\\
DeepSeek-V4.1-Flash & 60.17 & 57.33 & 5.80 & 75.77 & 72.22 & 58.73 & \recalldelta{-13.49}\\
Kimi-K3 & 60.00 & 57.33 & 3.13 & \textbf{77.10} & 67.46 & 54.76 & \recalldelta{-12.70}\\
Gemini-3.8-Flash & 63.67 & \textbf{60.00} & \textbf{1.57} & \textbf{79.22} & 77.78 & 60.32 & \recalldelta{-17.46}\\
GPT-5.6-sol & 46.33 & 40.50 & 1.72 & 69.39 & 61.90 & 34.13 & \recalldelta{-27.78}\\
\bottomrule\end{tabular*}\end{table}
\section{Experiments}
\subsection{Evaluation Setup}
We evaluate 24 configurations in three groups: ten visual detectors, six forensic specialist MLLMs, and eight general-purpose MLLMs (Table~\ref{tab:detectors}). Forensic specialist MLLMs use their model-specific inference instructions and greedy decoding. General-purpose MLLMs share an image-only prompt requesting a binary verdict and a short reason; it includes generation, editing, recreation and enhancement in the positive class while excluding ordinary cropping and compression alone. Temperature is zero, with thinking disabled where supported and set to low for GLM, Kimi and Gemini. Source links and evidence summaries are reserved for annotation and analysis.
\paragraph{Perception Encoder baselines.} PE-336 and PE-518 are binary image detectors built from the same pretrained Perception Encoder (PE-Core-L/14~\cite{perception_encoder}). We attach a two-layer linear head to the image representation and fine-tune the encoder and head with focal loss~\cite{focal_loss} on the training splits of WildFake, So-Fake-Set and WildRF~\cite{wildfake,so_fake,wildrf}. The suffix denotes the side length, in pixels, of the square input. We select each resolution's checkpoint by validation accuracy.
\begin{figure}[t]
\centering
\begin{minipage}{.42\columnwidth}\centering
\includegraphics[width=\linewidth]{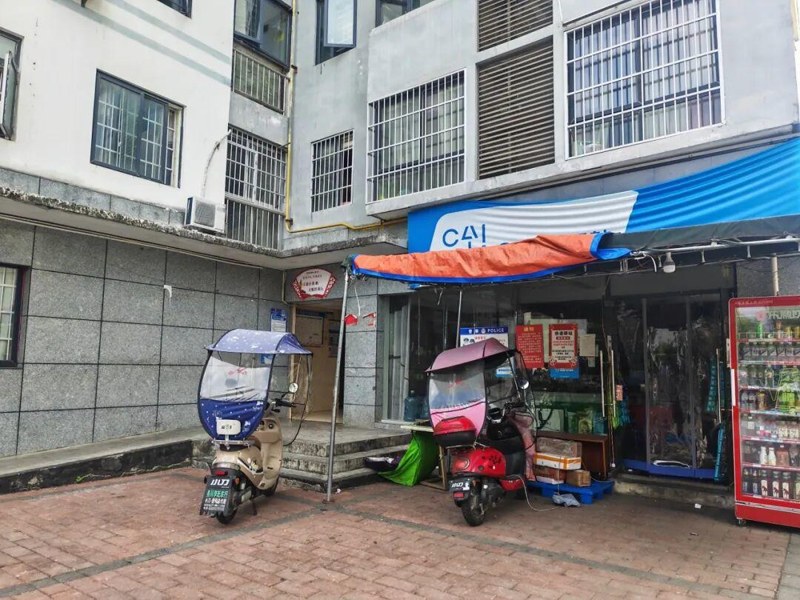}\\[-1pt]
\small (a) Original photograph
\end{minipage}\hfill
\begin{minipage}{.42\columnwidth}\centering
\includegraphics[width=\linewidth]{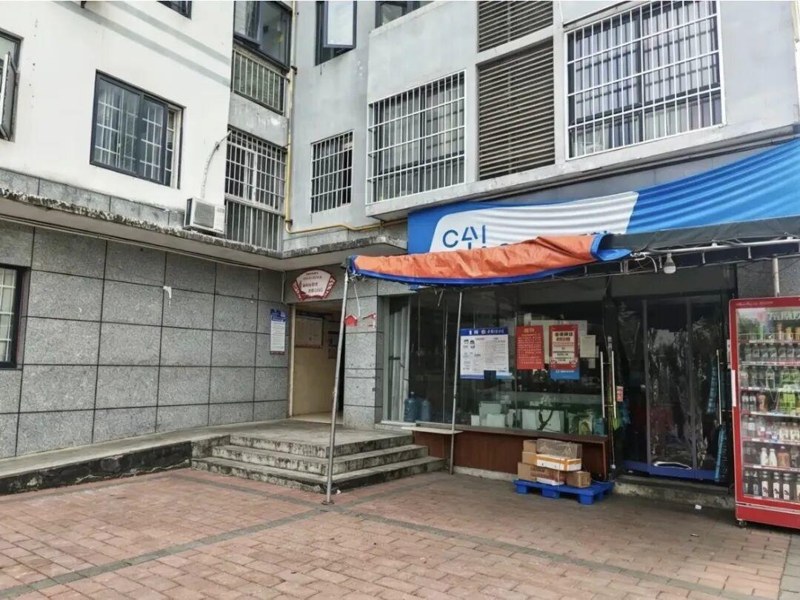}\\[-1pt]
\small (b) AI-edited image
\end{minipage}
\caption{Dazhong News reports that an edited photograph was submitted as evidence that a fire hazard had been cleared. Two parked vehicles visible in (a) disappear in (b).}\label{fig:editing-original}
\end{figure}
\paragraph{Scoring.} Visual detectors use a fake-score threshold of 0.5. For MLLMs, we extract the binary verdict from each response. We report AI recall, the false-positive rate (FPR) on photographs and balanced accuracy (BAcc), which weights the two classes equally. All configurations yield binary predictions on both views (Appendix H).

\subsection{Detection Performance and Transfer}
On processed images, mean BAcc across configurations is 73.14\% for general-purpose MLLMs, 60.44\% for forensic specialists and 58.01\% for visual detectors (Table~\ref{tab:detectors}). The general-purpose group has both the highest mean AI recall (51.85\%) and the lowest mean photograph FPR (5.56\%). The specialists' advantage over visual detectors is smaller: higher recall (38.67\% versus 27.62\%) comes with more false alarms (17.79\% versus 11.60\%). Individual methods overlap across groups. B-Free, for instance, exceeds five of the six specialists in BAcc. Gemini leads at 79.22\%, followed by Kimi at 77.10\%. Source-page weighting still places Gemini first (Appendix H).
General-purpose MLLMs can use event knowledge to assess a visually plausible scene. On the cropped Hollywood sign image, Kimi cites the fact that the sign survived the wildfire, along with irregularities in the letters and flames, and correctly flags the image. GPT accepts the same image, citing the lighting and terrain as consistent with a real fire (Appendix J).

\paragraph{Transfer across collections.} Table~\ref{tab:transfer} compares BAcc across benchmarks. For each visual detector, BAcc on WildAIGC-Bench falls below its mean across the eight external benchmarks with both classes. Relative to mean recall on the same eight benchmarks, AI recall drops by 21.18 to 69.87 percentage points across the ten visual detectors. Real-image accuracy improves for five of the ten. All ten visual detectors also lose AUROC, by 7.34 to 45.41 points relative to their means on the eight external benchmarks with both classes. Only EvalGEN is excluded from AUROC because it contains no real images. Thus the transfer loss extends to the ordering of images by fake score. PE-518 and PE-336 lead visual detectors on external mean BAcc (95.00\% and 92.34\%). On WildAIGC-Bench, increasing resolution raises recall from 20.50\% to 22.83\% and lowers FPR from 8.62\% to 1.72\%, yet PE-518 achieves only 60.55\% BAcc. Each visual detector also scores lower on WildAIGC-Bench than its own mean BAcc across Chameleon, SynthWildX, WildRF, and ITW-SM. Existing in-the-wild benchmarks underrepresent difficult circulating forgeries. The combination of photographic realism, localized AI edits, and degradation during online sharing in WildAIGC-Bench creates a more challenging distribution than standard generated imagery or selected online collections. All six forensic specialist MLLMs also lose BAcc, by 9.57--22.99 points relative to their external means. FakeVLM's higher recall comes with false alarms on nearly half the photographs.
\begin{table}[t]\centering\footnotesize
\caption{Balanced accuracy (\%). EvalGEN contains only AI images and reports recall. Gap is the mean BAcc across the eight two-class external benchmarks minus processed WildAIGC-Bench BAcc, in points. Bold marks the highest score within each family for each benchmark. Groups reflect evaluation focus; AIGIBench includes mixed sources, and RRD includes controlled recapture. Sample sizes are in Appendix E.}\label{tab:transfer}
\setlength{\tabcolsep}{2pt}\begin{tabular*}{\textwidth}{@{\extracolsep{\fill}}l*{11}{r}@{}}\toprule
 & \multicolumn{4}{c}{Standard} & \multicolumn{5}{c}{In-the-wild} & & \\[-2pt]\cmidrule(lr){2-5}\cmidrule(lr){6-10}
\raisebox{6pt}[0pt][0pt]{Method} & EvalGEN & Synthb. & UnivFD & AIGIB. & Cham. & SWX & WildRF & RRD & ITW-SM & \raisebox{1pt}[0pt][0pt]{\shortstack{WildAIGC-\\Bench}} & \raisebox{6pt}[0pt][0pt]{Gap}\\[-1pt]\midrule
\familyheading{12}{Visual detectors}\\
DRCT & 50.93 & 76.27 & 55.25 & 75.94 & 69.02 & 74.02 & 68.30 & 60.41 & 72.96 & 57.78 & 11.24\\
DDA & 96.85 & 92.13 & 84.73 & 84.99 & 82.41 & 90.72 & 87.74 & 71.16 & 87.23 & 64.60 & 20.54\\
AIDE & 56.26 & 79.38 & 79.13 & 81.13 & 62.85 & 67.93 & 67.83 & 57.44 & 62.34 & 54.48 & 15.28\\
B-Free & 91.28 & 96.97 & 86.70 & 84.83 & 76.59 & 95.63 & 92.80 & 71.92 & 89.73 & \textbf{66.60} & 20.30\\
Effort & 54.38 & 49.62 & 71.93 & 72.41 & 52.39 & 49.79 & 70.79 & 50.73 & 53.66 & 53.66 & 5.26\\
PGC & 57.19 & 80.48 & 87.68 & 87.01 & 56.10 & 79.32 & 86.08 & 58.17 & 75.07 & 50.46 & 25.78\\
OmniAID & 97.21 & 91.62 & 86.85 & 80.84 & \textbf{93.72} & 95.26 & 98.64 & 84.94 & \textbf{95.70} & 62.52 & 28.43\\
CO-SPY & 92.68 & 86.52 & 60.85 & 72.22 & 68.42 & 74.71 & 76.84 & 74.37 & 54.51 & 53.51 & 17.55\\
PE-336 & 99.96 & 98.22 & 94.39 & 93.60 & 87.68 & 94.26 & 97.89 & 88.56 & 84.15 & 55.94 & 36.40\\
PE-518 & \textbf{99.97} & \textbf{99.82} & \textbf{97.19} & \textbf{94.87} & 90.92 & \textbf{95.82} & \textbf{99.35} & \textbf{90.69} & 91.32 & 60.55 & 34.44\\
\midrule
\familyheading{12}{Forensic specialist MLLMs}\\
FakeShield & 96.10 & \textbf{97.50} & 52.15 & 70.62 & 50.13 & 69.66 & 77.20 & 75.55 & 61.60 & 59.29 & 10.01\\
FakeVLM & \textbf{99.00} & 61.48 & \textbf{75.35} & \textbf{81.01} & 74.55 & 73.80 & 74.16 & 65.55 & 71.23 & 58.49 & 13.66\\
UniGenDet & 70.60 & 83.33 & 63.05 & 68.47 & 76.71 & 81.77 & 86.19 & 73.45 & 85.34 & \textbf{67.72} & 9.57\\
\shortstack[l]{BusterX++\\{\scriptsize [2026 release version]}} & 77.80 & 80.58 & 62.50 & 71.99 & 63.28 & \textbf{85.47} & 90.87 & 62.80 & \textbf{88.13} & 55.28 & 20.42\\
DeepVRM & 96.55 & 71.45 & 66.95 & 74.86 & \textbf{91.30} & 82.35 & \textbf{91.48} & 74.50 & 77.06 & 55.76 & 22.99\\
Veritas++ & 98.90 & 82.85 & 64.30 & 77.99 & 86.33 & 85.13 & 89.74 & \textbf{76.25} & 81.95 & 66.10 & 14.47\\
\bottomrule\end{tabular*}
\end{table}
\subsection{Shared Misses and Complementarity}
\begin{figure}[t]\centering
\includegraphics[width=\textwidth]{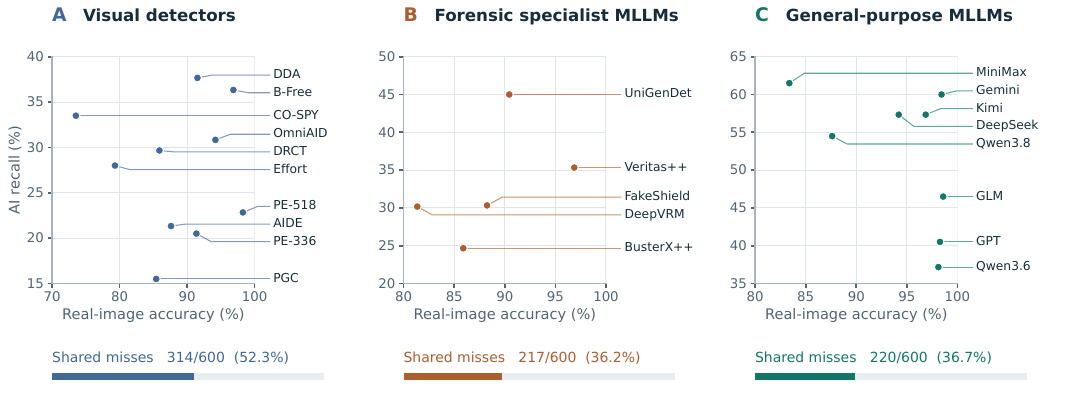}
\caption{Detection performance on processed images. Each point is one model. Toward the upper right, a model detects more AI images and correctly accepts more photographs. Axis ranges vary by panel. Panel B uses BusterX++ [2026 release version]. FakeVLM (50.47\% real-image accuracy, 66.50\% AI recall) is omitted from the scatter plot. The bars count AI images missed by at least three quarters of each full family, including FakeVLM in the specialist group. Of the 600 AI images, 96 meet the threshold in all three families.}\label{fig:operating-points}
\end{figure}
\begin{figure}[t]
\centering\includegraphics[width=\textwidth]{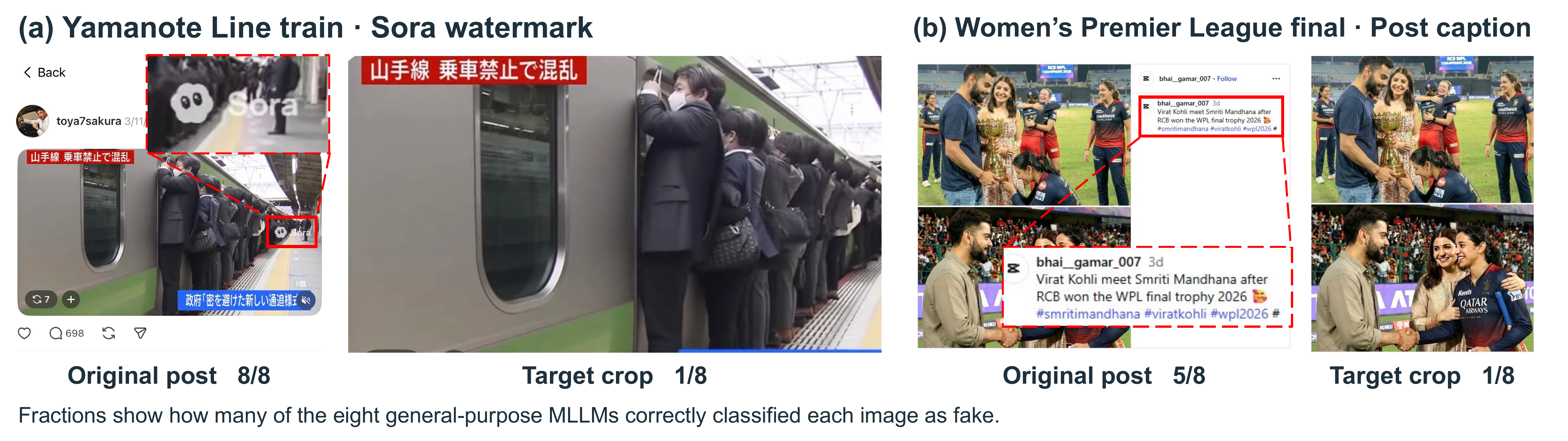}
\caption{General-purpose MLLM predictions before and after cropping. In (a), cropping removes a Sora watermark. In (b), both pictures remain, but the Instagram caption placing the celebration at the 2026 final is removed. Sources: HKBU Fact Check and Newschecker.}\label{fig:context-crop}
\end{figure}
On 222 of the 600 processed AI images (37.00\%), at least three quarters of all 24 configurations predict real. On 96 images (16.00\% of the 600), at least three quarters of the models in \emph{each} of the three families predict real. Figure~\ref{fig:operating-points} reports the number of images meeting the threshold within each family. The within-family thresholds are 8 of 10 visual detectors, 5 of 6 forensic specialist MLLMs and 6 of 8 general-purpose MLLMs. The respective thresholds are met on 314, 217 and 220 images. Shared misses span full-scene generation and edits, including a fabricated tower near Taipei 101 and a collage placing a football official at two matches. Of the 96 shared misses, 84 occur among the 474 AI images with identical raw and processed files, and 12 among the 126 with distinct views.

\paragraph{Complementary errors.} Different evidence sources motivate testing whether detector families recover one another's misses. Of 94 AI images missed by all visual detectors, Gemini detects 45 and Kimi 32. Conversely, B-Free detects 79 Gemini misses. We evaluate all cross-family model pairs, including forensic specialists, and flag an image when either member predicts fake. BAcc improves in 35/80 visual--general, 24/48 specialist--general and 44/60 visual--specialist pairs, relative to the general-purpose or specialist MLLM alone. Complementary detections improve some combinations, but additional false alarms limit simple decision fusion. Appendix I reports recall and FPR changes.

\subsection{Paired Image Comparisons}
Across the 126 AI images with distinct acquired and cropped views, cropping raises mean recall by 5.00 percentage points for visual detectors and 4.63 points for forensic specialist MLLMs, but lowers it by 12.80 points for general-purpose MLLMs. Eight of ten visual detectors gain recall, while seven of eight general-purpose MLLMs lose recall (Table~\ref{tab:detectors}). The largest drop for general-purpose MLLMs occurs on the 12 posts where cropping removes both a generation cue and text describing the claim. Mean recall falls by 44.79 points on these posts and rises by 1.07 points on the 35 where neither cue is identified in the removed region (Appendix C).

Figure~\ref{fig:context-crop} shows two sources of contextual evidence. GPT cites the Sora watermark to reject the train post, but accepts the crop as plausible news footage. For the cricket post, it questions whether the players appeared together at the final claimed in the caption. Without that text, it accepts the same two pictures based on their lighting and interactions. Watermarks disclose generation directly; social captions connect visual content to event knowledge. These cases show how general-purpose MLLMs use both generation disclosures and event knowledge to detect fabrications that appear visually plausible.

\paragraph{Distinguishing edits from originals.} A pair is correct only when the detector accepts the original photograph and rejects its AI edit. On 17 of the 20 pairs, a majority of configurations accept both images. In the storefront example (Fig.~\ref{fig:editing-original}), removing the vehicles leaves the scene convincing enough that no configuration correctly distinguishes the two versions. These results expose a shared failure to detect localized changes that preserve the photographic appearance of the surrounding scene. Appendix C reports joint decisions and score comparisons.

\section{Conclusion}
WildAIGC-Bench uses recurring collection and human verification to benchmark detection in online circulation. It reveals a substantial transfer gap in balanced accuracy and score ranking, with shared misses spanning realistic fabrications and localized edits. Paired comparisons show how visual detectors benefit from cleaner target regions, while general-purpose MLLMs also exploit contextual cues and event knowledge.

The collection remains concentrated in some sources, and prior exposure to circulated images may influence model performance. Future releases will cover more sources and add verified samples and edit--original pairs. Future research should improve localized manipulation detection and robustness to online dissemination, integrating visual evidence with verifiable event knowledge at low false-positive rates.
\section*{LLM Usage Statement}
We used GPT, Gemini, and Qwen to search for candidate images. GPT then screened the candidate images and assigned preliminary labels. Human reviewers verified each label against the source evidence and decided which images to include. The authors used OpenAI Codex to help implement the evaluation, categorize image topics, design figures, organize references, and identify cases meeting author-defined criteria for analysis. Codex provided preliminary observations on the selected cases. The authors used Codex to draft and revise the text based on an outline they had prepared.
\clearpage\newgeometry{left=.75in,right=.75in,top=.7in,bottom=.75in}\small\appendix\section*{Supplementary Material}\setcounter{figure}{0}\setcounter{table}{0}\renewcommand{\theHfigure}{app.\arabic{figure}}\renewcommand{\theHtable}{app.\arabic{table}}
\renewcommand{\thetable}{A\arabic{table}}
\renewcommand{\thefigure}{A\arabic{figure}}
\noindent\textbf{Contents.} Sections~\ref{sec:app-construction}--\ref{sec:app-data} document construction and dataset composition. Section~\ref{sec:app-crops} analyzes crops, editing originals and same-event pairs; Section~\ref{sec:app-atlas} illustrates detector misses. Sections~\ref{sec:app-coverage}--\ref{sec:app-external} report external coverage and results. Sections~\ref{sec:app-config}--\ref{sec:app-full} provide inference settings, the exact prompt and benchmark diagnostics. Section~\ref{sec:app-shared} examines shared errors and cross-family complementarity. Section~\ref{sec:app-context} presents contextual and event-knowledge cases.

\section{Construction Workflow and Annotation Details}
\label{sec:app-construction}
The construction pipeline assigns collection, initial screening, and preliminary annotation to scheduled automatic tasks that are already running. Human reviewers then verify whether the automatic annotations are correct and perform quality control to decide whether each sample enters the benchmark. Automatic annotations remain provisional until this review is complete.

\textbf{Image-specific evidence.} To verify an automatic annotation, reviewers check that the supporting evidence concerns the target image and establishes AI generation or editing. A false event claim alone is insufficient: a genuine photograph may also be shared with an incorrect description. Creator disclosures, visible platform watermarks, and explicit fact-checking findings can support admission. Invisible-watermark findings are taken from publicly reported fact-checking results rather than independent tool runs by the team. Candidates with insufficient evidence or uncertain AI involvement are excluded.

\textbf{Quality control and image preparation.} Human reviewers make the final inclusion decision, excluding obviously stylized artwork and unusable images while retaining artifacts introduced during online circulation. Accepted samples preserve their acquired image and supporting source information. A target-region crop removes peripheral text and interface elements where appropriate, while the raw image remains available for paired evaluation. No AI-image detectors are used for sample selection.

The workflow is illustrated in Fig.~2 of the main paper. The details here clarify the evidence used for admission and the distinction between preliminary annotation and final verification.
\section{Dataset Composition and Record Structure}\label{sec:app-data}
The benchmark pairs circulating AI-generated or edited images with genuine photographs of related subjects. The public release will include the images, a sample index and image-level annotations. The two classes serve complementary evaluation roles: AI images measure detection sensitivity, while genuine photographs measure false alarms.
\begin{center}\small
\captionof{table}{Collection statistics. Pages are recorded source URLs, and hostnames identify publishing websites.}\label{tab:app-composition}
\begin{tabular}{lrr}\toprule
Property & AI images & Photographs\\\midrule
Images & 600 & 638\\
Primary source pages & 487 & 89\\
Source hostnames & 162 & 34\\
Identical raw/processed inputs & 474 & 638\\
Distinct raw/processed inputs & 126 & 0\\
\bottomrule\end{tabular}\end{center}
\subsection{Image Records and Linked References}
Each record associates the acquired image and processed view with a source page, a description of the depicted event or claim, and a summary of the evidence supporting its label. Additional references preserve disclosures, investigations and credited originals when available. Evidence describes the particular target, including which part was generated, edited or reconstructed when the source establishes this. Generator identity is not inferred from visual style alone.

An editing-original relation links an AI modification to its documented source photograph: 20 edits are linked to 18 originals. A same-event relation links a fabrication to an authentic photograph of the corresponding event, without implying that one was edited from the other: 24 links connect 20 AI images and 21 photographs. These relation types support different questions, and their evaluation is reported separately in Section~\ref{sec:app-paired}.
\subsection{Source Coverage}
The collection spans many websites, with uneven numbers of images per source. Galleries can contribute multiple photographs, whereas a fact-check page often centers on one claim. The leading sources below make this composition explicit. Section~\ref{sec:app-full} compares image-weighted performance with source-page-weighted performance.
\begin{center}\small
\captionof{table}{The five largest source hostnames in each class.}\label{tab:app-hosts}
\begin{tabular}{lrlr}\toprule
AI-image source & Images & Photograph source & Images\\\midrule
leadstories.com & 131 & www.abc.net.au & 201\\
www.snopes.com & 49 & abcnews.com & 177\\
factcheck.afp.com & 43 & www.cbsnews.com & 80\\
maldita.es & 30 & time.com & 63\\
fullfact.org & 23 & www.theguardian.com & 44\\
\bottomrule\end{tabular}\end{center}
\subsection{Continued Collection and Label Maintenance}
Recurring retrieval follows new reports and disclosures, while human verification remains the admission step for every new image. Broad coverage comes from searching across complementary source types; timely updates come from repeated discovery; label reliability comes from checking image-specific evidence before inclusion. These properties concern the collected dataset, with automation supporting the efficiency and continuity of construction. Retaining both the evidence and the acquired image also makes later reinspection possible when a source publishes a clarification or additional provenance becomes available.

\paragraph{Sources for the main montage.} Sources (linked by panel): (a) \href{https://maldita.es/desinfo/20261001/imagen-del-interior-de-la-linea-3-de-metro-de-madrid-en-la-zona-universitaria-inundada/}{Maldita.es}; (b) \href{https://crhoy.com/entretenimiento/no-caiga-fotos-que-circulan-de-la-boda-de-cristiano-y-georgina-son-ia/}{CRHoy}; (c) \href{https://factcheck.afp.com/doc.afp.com.C98N6CQ}{AFP}; (d) \href{https://hmgroup.com/news/hm-continues-its-exploration-of-creativity-with-ai/}{H\&M}; (e) \href{https://www.unexplained-mysteries.com/news/368350/alleged-1883-bigfoot-photograph-has-been-exposed-as-a-hoax}{Unexplained Mysteries}; (f) \href{https://www.snopes.com/fact-check/k9-valor-medal-honor-story/}{Snopes}; (g) \href{https://english.factcrescendo.com/2025/05/28/is-ben-jerrys-selling-a-free-palestine-ice-cream/}{Fact Crescendo}; (h) \href{https://www.4anet.com/p/01m4csjyqm986pe0gzfwxsfa9d}{4A Advertising}; (i) \href{https://leadstories.com/hoax-alert/2026/02/fact-check-image-of-buildings-on-fire-is-not-real-photo-of-puerto-vallarta.html}{Lead Stories}; (j) \href{https://fullfact.org/politics/edited-image-reform-uk-sarah-pochin-campaign/}{Full Fact}.

\clearpage
\section{Paired Image Analyses}\label{sec:app-crops}\label{sec:app-paired}
We compare three types of paired inputs: acquired images and their target crops, AI edits and their documented originals, and fabricated and genuine images of the same event. The first comparison retains the AI label in both views; the latter two require correct decisions on both classes.
\subsection{Acquired Images and Target Crops}
Of the 600 AI images, 126 have distinct acquired and cropped views. Table~\ref{tab:removed-cues} groups these pairs by the evidence outside the target region. Figure~6 of the main paper illustrates two such cues: a generator watermark and a caption identifying the claimed event.

\paragraph{Cues removed by cropping.} For each of the 126 pairs, we masked the retained rectangle in the acquired image to isolate the removed content. An author verified the cue labels for all pairs. Generation cues include visible generator marks, AI disclosures and AI-generation tags. Claim text identifies or describes the depicted subject, event or product. Account names, timestamps and interface controls alone do not establish either cue. We grouped images by the two annotations and kept ambiguous margins in a separate row. Group differences describe the observed crops, which can also change target scale.

\begin{center}\small
\captionof{table}{Paired recall changes by removed cue. Gains and losses count image--model decisions across the eight general-purpose MLLMs. The final three columns give family mean changes in recall (processed minus raw, percentage points).}\label{tab:removed-cues}
\begin{tabular}{lrrrrrr}\toprule
Removed cue & Images & Gains & Losses & General & Specialist & Visual\\\midrule
Generation cue only & 25 & 1 & 47 & -23.00 & -2.00 & +3.60\\
Claim text only & 44 & 21 & 59 & -10.80 & +12.12 & +5.23\\
Both cues & 12 & 1 & 44 & \textbf{-44.79} & -6.94 & +5.83\\
Neither cue identified & 35 & 21 & 18 & +1.07 & +2.86 & +4.57\\
Unresolved & 10 & 5 & 10 & -6.25 & +8.33 & +8.00\\
\midrule
\textbf{All} & 126 & 49 & 178 & \textbf{-12.80} & \textbf{+4.63} & \textbf{+5.00}\\
\bottomrule\end{tabular}\end{center}
Across the eight general-purpose MLLMs, the 12 images with both cues account for 44 lost detections and one gain. On the 35 images with neither cue identified, gains and losses nearly balance (21 versus 18). Visual detectors gain recall in every group.

\paragraph{Holding the target at its acquired scale.} On all 126 changed images, we also blur only the pixels outside the recorded crop box, copy the untouched target pixels back into the frame, and score the full image at its original dimensions. The Gaussian radius is the larger of eight pixels and one twenty-fifth of the longer side (rounded). With the same checkpoints and thresholds, PE-518 detects 13 AI images in the raw condition, 16 with blurred margins and 18 after cropping. DDA detects 41, 43 and 43, respectively. For DDA, the last two conditions nevertheless disagree on 16 images, split evenly between directions. Similar overall recall thus conceals sensitivity to the framing of particular targets. The blur introduces an artificial boundary; this intervention measures sensitivity to peripheral detail at fixed input geometry.

\clearpage
\subsection{Decision Changes Across Paired Inputs}
The 474 AI images and 638 photographs with identical acquired and processed files share one prediction per model. Crop gains and losses therefore come from the 126 distinct pairs.
Table~\ref{tab:all} summarizes all three paired comparisons. A gain means the raw image was missed and the crop was detected. A loss means the opposite. The editing-original and same-event columns count pairs for which both predictions are correct. Counts are reported per recorded relation; an image can participate in more than one relation.
\begin{table}[ht!]\centering\small\caption{Paired diagnostics for every evaluated configuration.}\label{tab:all}
\begin{tabular}{lrrrr}\toprule Model & Crop gains & Crop losses & \shortstack{Editing originals\\Both correct / 20} & \shortstack{Same-event photos\\Both correct / 24}\\\midrule
\familyheading{5}{Visual detectors}\\
DRCT & 11 & 12 & 4 & 2\\
AIDE & 33 & 4 & 2 & 3\\
B-Free & 8 & 5 & 4 & 8\\
CO-SPY & 17 & 5 & 3 & 5\\
Effort & 13 & 9 & 3 & 5\\
DDA & 10 & 8 & 5 & 8\\
OmniAID & 5 & 0 & 1 & 2\\
PGC & 4 & 5 & 1 & 0\\
PE-336 & 8 & 3 & 2 & 5\\
PE-518 & 7 & 2 & 0 & 6\\
\familyheading{5}{Forensic specialist MLLMs}\\
FakeShield & 13 & 17 & 3 & 4\\
FakeVLM & 22 & 10 & 6 & 6\\
UniGenDet & 23 & 11 & 5 & 8\\
BusterX++ & 14 & 9 & 2 & 6\\
DeepVRM & 12 & 12 & 4 & 4\\
Veritas++ & 16 & 6 & 4 & 8\\
\familyheading{5}{General-purpose MLLMs}\\
Qwen3.8-27B & 14 & 20 & 1 & 12\\
Qwen3.6-35B-A3B & 4 & 16 & 3 & 8\\
GLM-5.3-Flash & 5 & 26 & 2 & 10\\
MiniMax-M3 & 15 & 15 & 4 & 12\\
DeepSeek-V4.1-Flash & 5 & 22 & 6 & 10\\
Kimi-K3 & 5 & 21 & 7 & 10\\
Gemini-3.8-Flash & 0 & 22 & 8 & 12\\
GPT-5.6-sol & 1 & 36 & 1 & 6\\
\bottomrule\end{tabular}\end{table}

\clearpage
\subsection{Edits and Their Originals}\label{sec:app-ranking}
Figure~4 of the main paper shows the storefront photograph and its vehicle-removal edit. An editing-original link permits a within-scene score comparison. A rank win requires a strictly higher fake score for the edit. Table~\ref{tab:edit-rank} also counts pairs for which both decisions are correct.\begin{center}\small\captionof{table}{Score ordering and joint decisions on 20 edit--original pairs.}\label{tab:edit-rank}\begin{tabular}{lrr}\toprule Model & Higher score on edit (/20) & Both correct (/20)\\\midrule AIDE & 8 & 2\\
B-Free & 13 & 4\\
CO-SPY & 14 & 3\\

DDA & 14 & 5\\
DRCT & 14 & 4\\
Effort & 11 & 3\\

OmniAID & 12 & 1\\
PE-336 & 10 & 2\\
PE-518 & 9 & 0\\
PGC & 9 & 1\\
\bottomrule\end{tabular}\end{center}\paragraph{Why ranking adds information.} B-Free scores the edit higher on 13 pairs but classifies both images correctly on only four. DDA, DRCT and CO-SPY each rank 14 edits above their originals. A threshold can separate a correctly ordered pair, although the required threshold can differ across pairs. If the edit scores at or below the original, no threshold can classify both correctly.
\paragraph{Joint decisions.} Accepting both images misses the edit, rejecting both raises a false alarm on the original, and rejecting only the original misclassifies both. The 20 edits are linked to 18 distinct original photographs.

\subsection{Same-Event Comparisons}
The 24 same-event links connect fabrications and genuine photographs of a corresponding event, without requiring that one image was edited from the other. A pair is correct only when the fabrication is rejected and the genuine photograph is accepted. This comparison tests authenticity discrimination under a shared event context, whereas editing-original pairs isolate changes within the same source scene. Table~\ref{tab:all} reports both criteria separately. Gemini correctly classifies both members in 12 of the 24 same-event pairs and eight of the 20 editing-original pairs. Strong aggregate performance therefore leaves substantial room to distinguish individual fabrications from related authentic photographs.

\clearpage
\section{Examples of Detector Misses}\label{sec:app-atlas}
The six examples illustrate different forms of modification across public affairs, commercial imagery, infrastructure and eyewitness reports. For each example in Fig.~\ref{fig:misses}, the count gives the number of evaluated configurations that classify the processed AI image as real. Section~\ref{sec:app-shared} reports the frequency of shared misses across the full AI class.

\begingroup

\noindent
\begin{minipage}[t]{0.315\linewidth}\centering
  \begingroup\setlength{\fboxsep}{0pt}%
  \colorbox{black!5}{\vbox to 1.82in{\vfil
    \hbox to \linewidth{\hfil\includegraphics[width=\linewidth,height=1.82in,keepaspectratio]{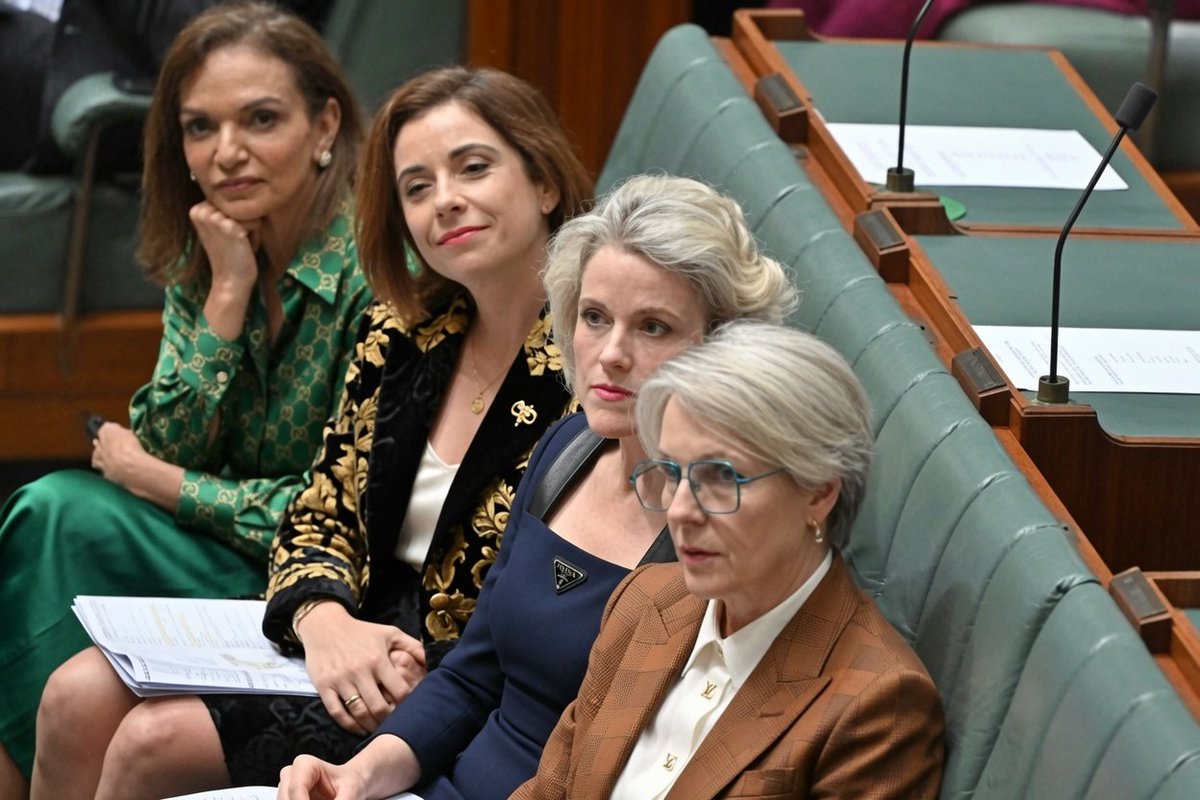}\hfil}%
    \vfil}}\endgroup\par\smallskip
\small\textbf{(a) Clothing edits}\par 22 of 24 predict real\par\smallskip\raggedright Luxury-brand clothing and logos were added to a photograph of Australian ministers, as shown by comparison with the original press photograph. Source: \href{https://www.aap.com.au/factcheck/photo-of-labor-ministers-wearing-high-end-fashion-is-an-ai-counterfeit/}{AAP FactCheck}.\end{minipage}
\hfill
\begin{minipage}[t]{0.315\linewidth}\centering
  \begingroup\setlength{\fboxsep}{0pt}%
  \colorbox{black!5}{\vbox to 1.82in{\vfil
    \hbox to \linewidth{\hfil\includegraphics[width=\linewidth,height=1.82in,keepaspectratio]{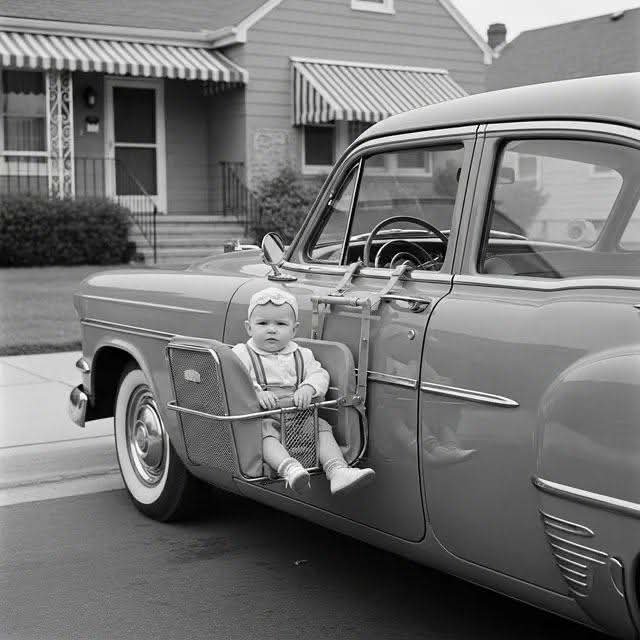}\hfil}%
    \vfil}}\endgroup\par\smallskip
\small\textbf{(b) Fabricated historical image}\par 15 of 24 predict real\par\smallskip\raggedright A black-and-white image of a baby seat outside a car circulated as a historical photograph. Snopes reports a positive SynthID check. Source: \href{https://www.snopes.com/fact-check/image-baby-car-seat-outside/}{Snopes}.\end{minipage}
\hfill
\begin{minipage}[t]{0.315\linewidth}\centering
  \begingroup\setlength{\fboxsep}{0pt}%
  \colorbox{black!5}{\vbox to 1.82in{\vfil
    \hbox to \linewidth{\hfil\includegraphics[width=\linewidth,height=1.82in,keepaspectratio]{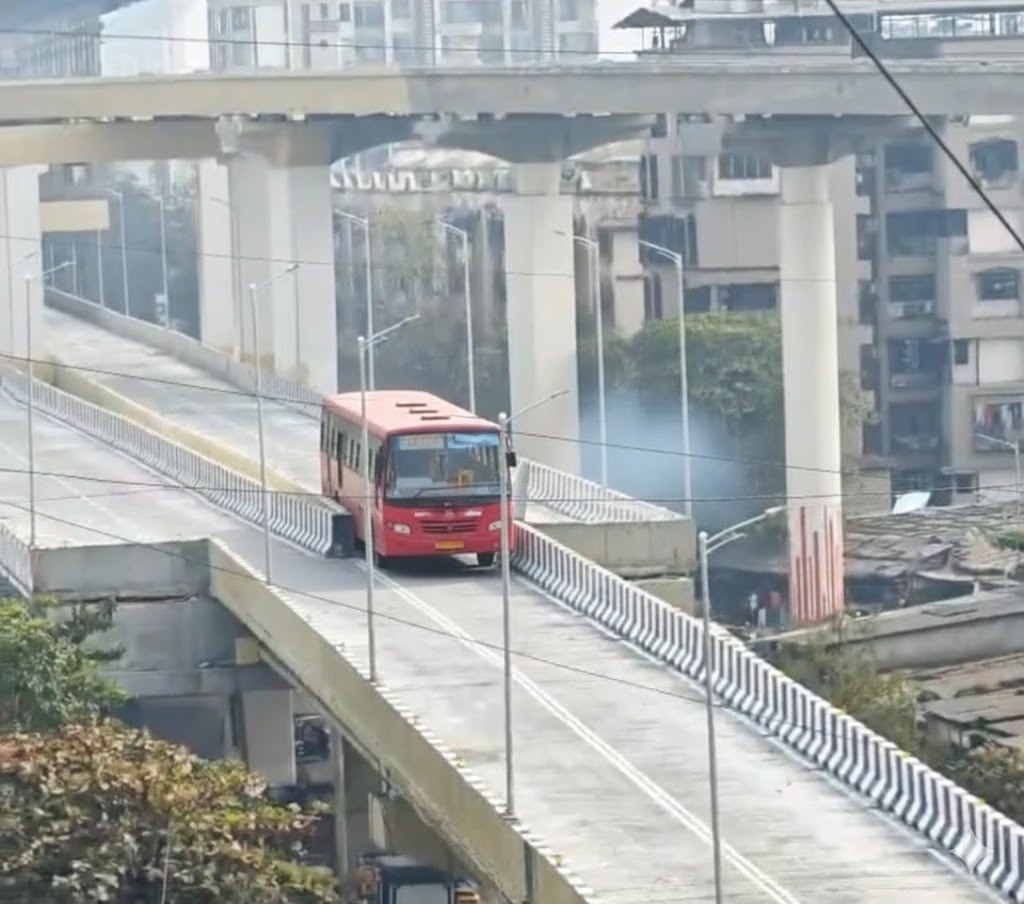}\hfil}%
    \vfil}}\endgroup\par\smallskip
\small\textbf{(c) Object insertion}\par 20 of 24 predict real\par\smallskip\raggedright A bus was added to a real Mumbai flyover scene, creating the appearance of a vehicle stuck where the lanes narrow. Source: \href{https://www.altnews.in/bus-stuck-at-narrowing-of-lanes-on-mumbai-flyover-no-photo-shared-by-kerala-congress-is-ai-generated/}{Alt News}.\end{minipage}
\par\bigskip\noindent
\begin{minipage}[t]{0.315\linewidth}\centering
  \begingroup\setlength{\fboxsep}{0pt}%
  \colorbox{black!5}{\vbox to 1.82in{\vfil
    \hbox to \linewidth{\hfil\includegraphics[width=\linewidth,height=1.82in,keepaspectratio]{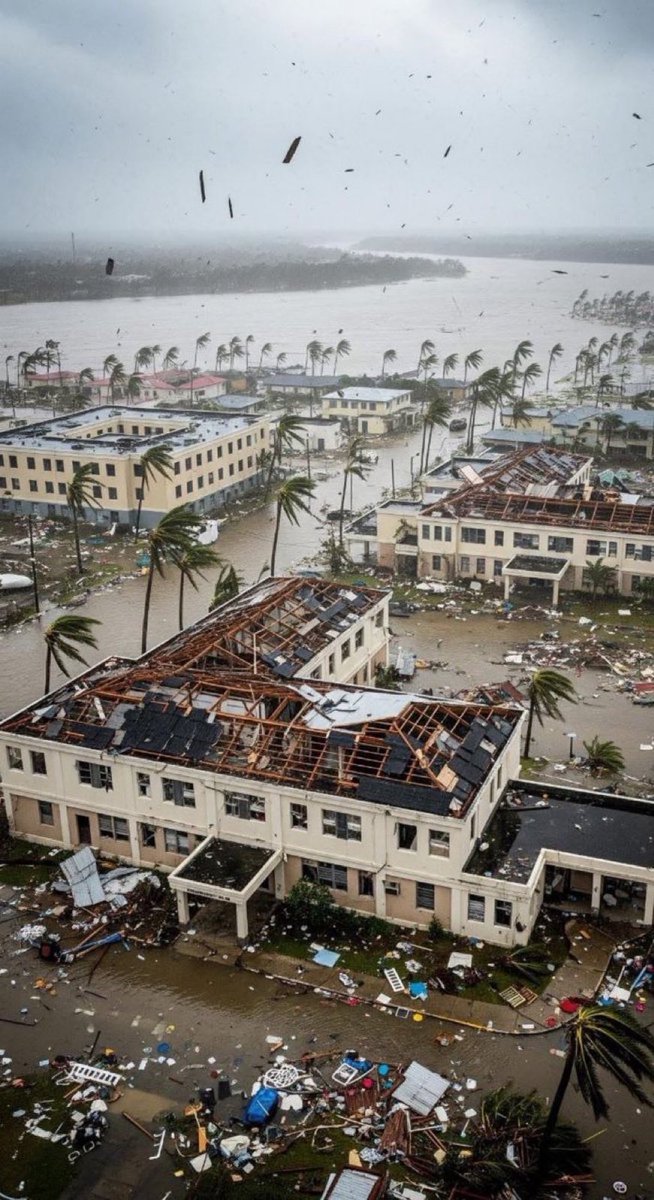}\hfil}%
    \vfil}}\endgroup\par\smallskip
\small\textbf{(d) Fabricated disaster scene}\par 16 of 24 predict real\par\smallskip\raggedright The purported aerial view of a hurricane-damaged Jamaican hospital differs from its actual layout. Full Fact reports confirmation of a SynthID watermark. Source: \href{https://fullfact.org/environment/jamaica-hurricane-melissa-photo-fake/}{Full Fact}.\end{minipage}
\hfill
\begin{minipage}[t]{0.315\linewidth}\centering
  \begingroup\setlength{\fboxsep}{0pt}%
  \colorbox{black!5}{\vbox to 1.82in{\vfil
    \hbox to \linewidth{\hfil\includegraphics[width=\linewidth,height=1.82in,keepaspectratio]{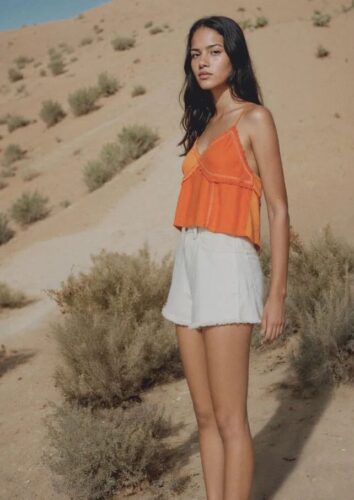}\hfil}%
    \vfil}}\endgroup\par\smallskip
\small\textbf{(e) Generative fashion imagery}\par 16 of 24 predict real\par\smallskip\raggedright Mango's disclosed generative-AI campaign depicts a model in a desert setting, combining photographic advertising style with generated imagery. Source: \href{https://invidis.com/news/2024/07/mango-entire-campaign-genai-created/}{invidis}.\end{minipage}
\hfill
\begin{minipage}[t]{0.315\linewidth}\centering
  \begingroup\setlength{\fboxsep}{0pt}%
  \colorbox{black!5}{\vbox to 1.82in{\vfil
    \hbox to \linewidth{\hfil\includegraphics[width=\linewidth,height=1.82in,keepaspectratio]{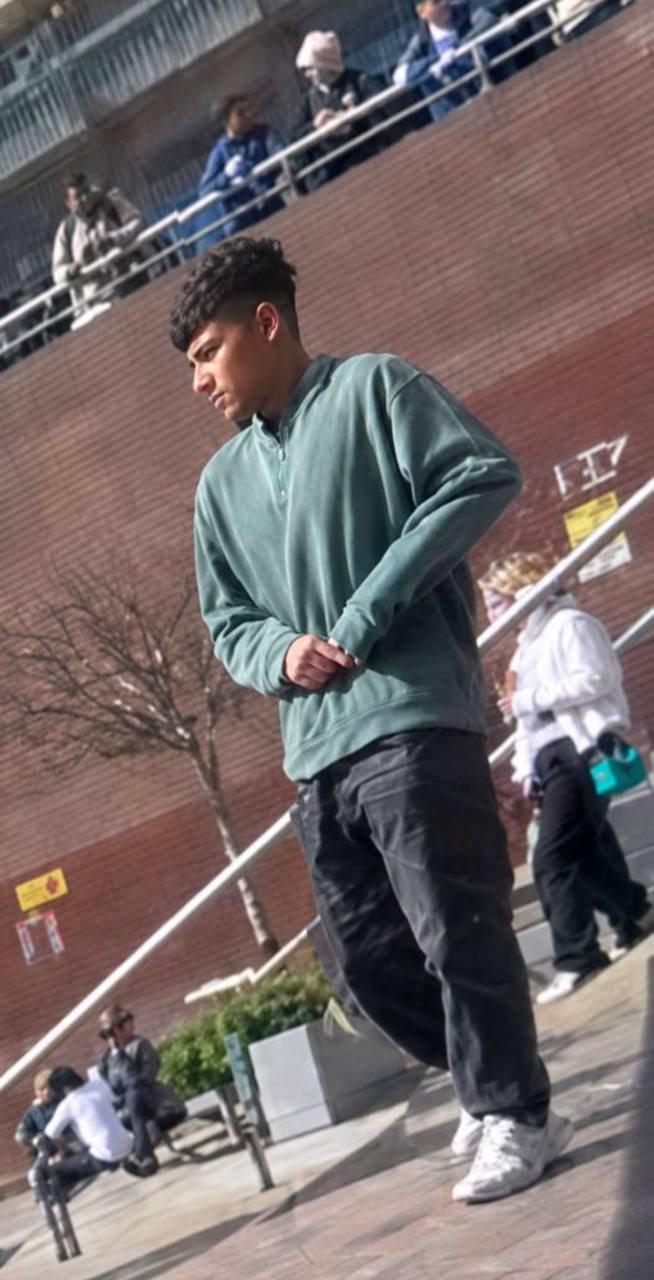}\hfil}%
    \vfil}}\endgroup\par\smallskip
\small\textbf{(f) Reconstructed details}\par 23 of 24 predict real\par\smallskip\raggedright AI reconstruction of low-resolution footage introduces clear facial and clothing details that the original recording does not establish. Source: \href{https://leadstories.com/hoax-alert/2026/08/fact-check-viral-images-are-not-photos-or-screenshots-showing-real-face-of-warsaw-attack-suspect.html}{Lead Stories}.\end{minipage}
\par\bigskip
\endgroup
\captionof{figure}{Examples of detector misses across public affairs, historical imagery, infrastructure, disasters, advertising, and eyewitness-style images. Misses occur across both modified photographs and generated content, rather than a single visual genre.}\label{fig:misses}
\paragraph{Patterns across cases.} A photograph with altered clothing preserves most scene content. A reconstructed face can introduce a misleading identity while retaining its apparent surroundings. These two tasks can be difficult even when the delivered image is clear. The published fashion campaign illustrates a different source of difficulty: professionally produced imagery may resemble conventional advertising photography without making any factual claim about a news event.

\clearpage
\section{External Evaluation Coverage}\label{sec:app-coverage}
Visual detectors use the complete retained test lists. Forensic specialist MLLMs use fixed lists shared across the newly evaluated specialists. Five collections are restricted to 2,000 images each; Synthbuster, SynthWildX, WildRF and ITW-SM use their complete retained lists. Both classes are evaluated whenever present. General-purpose MLLMs are evaluated on WildAIGC-Bench and are not included in the external comparison.
\begin{center}\small
\captionof{table}{External evaluation sizes. The specialist lists contain 27,615 images in total.}\label{tab:app-coverage}
\begin{tabular}{lrrrr}\toprule
Benchmark & Visual: AI & Visual: real & Specialist: AI & Specialist: real\\\midrule
EvalGEN & 55,285 & 0 & 2,000 & 0\\
Synthbuster & 3,000 & 1,000 & 3,000 & 1,000\\
UnivFD & 45,160 & 45,169 & 1,000 & 1,000\\
AIGIBench & 105,298 & 105,500 & 999 & 1,001\\
Chameleon & 11,170 & 14,863 & 858 & 1,142\\
SynthWildX & 1,354 & 427 & 1,354 & 427\\
WildRF & 925 & 909 & 925 & 909\\
RRDataset & 25,500 & 25,491 & 1,000 & 1,000\\
ITW-SM & 5,000 & 5,000 & 5,000 & 5,000\\
\bottomrule\end{tabular}\end{center}
\subsection{AI-Image Recall}
Table~\ref{tab:app-recall} complements the BAcc comparison in the main paper by isolating sensitivity to AI images. EvalGEN contributes recall only because it has no real-image class. Means below average over benchmarks, rather than pooling their different numbers of images.
\begin{center}\small
\captionof{table}{AI-image recall (percent). Mean (9) averages the nine benchmarks equally, including EvalGEN. The recall drop in the main paper uses the eight benchmarks with both classes.}\label{tab:app-recall}
\setlength{\tabcolsep}{3.5pt}\begin{tabular}{lrrrrrrrrrr}\toprule
Model & EvalGEN & Synthb. & UnivFD & AIGIB. & Cham. & SWX & WildRF & RRD & ITW-SM & Mean (9)\\\midrule
\familyheading{11}{Visual detectors}\\
DRCT & 50.93 & 55.93 & 21.96 & 56.50 & 58.18 & 59.75 & 58.38 & 36.27 & 59.82 & 50.86\\
AIDE & 56.26 & 65.27 & 81.01 & 70.90 & 33.98 & 48.74 & 55.14 & 22.38 & 43.08 & 52.97\\
B-Free & 91.28 & 94.23 & 82.66 & 71.89 & 62.01 & 98.52 & 93.19 & 47.86 & 88.86 & 81.17\\
CO-SPY & 92.68 & 75.53 & 42.44 & 61.62 & 52.19 & 73.78 & 75.68 & 64.33 & 51.82 & 65.56\\
Effort & 54.38 & 97.03 & 91.20 & 83.26 & 51.91 & 34.71 & 69.30 & 15.10 & 37.26 & 59.35\\
DDA & 96.85 & 97.57 & 83.59 & 74.45 & 70.78 & 92.91 & 87.24 & 47.00 & 84.54 & 81.66\\
OmniAID & 97.21 & 83.43 & 80.77 & 71.15 & 90.86 & 93.80 & 98.59 & 72.87 & 95.04 & 87.08\\
PGC & 57.19 & 81.07 & 91.99 & 81.00 & 16.03 & 71.05 & 88.43 & 27.45 & 64.92 & 64.35\\
PE-336 & 99.96 & 96.53 & 99.45 & 97.00 & 76.06 & 90.62 & 96.54 & 77.97 & 80.28 & 90.49\\
PE-518 & 99.97 & 99.73 & 99.42 & 96.09 & 82.16 & 94.68 & 98.81 & 82.29 & 88.40 & 93.51\\
\midrule
\familyheading{11}{Forensic specialist MLLMs}\\
FakeShield & 96.10 & 98.90 & 64.20 & 89.69 & 19.35 & 48.23 & 65.41 & 88.60 & 51.86 & 69.15\\
FakeVLM & 99.00 & 95.87 & 92.70 & 94.09 & 99.53 & 94.90 & 92.76 & 58.70 & 88.10 & 90.63\\
UniGenDet & 70.60 & 67.17 & 66.30 & 88.49 & 95.80 & 86.26 & 88.32 & 69.80 & 89.30 & 80.23\\
BusterX++ & 77.80 & 67.27 & 61.80 & 63.16 & 66.67 & 90.62 & 93.30 & 56.30 & 86.96 & 73.76\\
DeepVRM & 96.55 & 97.20 & 88.00 & 84.28 & 83.22 & 75.48 & 92.76 & 58.40 & 78.44 & 83.81\\
Veritas++ & 98.90 & 83.00 & 44.30 & 69.77 & 93.24 & 93.21 & 97.41 & 70.50 & 95.40 & 82.86\\
\bottomrule\end{tabular}\end{center}
\paragraph{Complete specialist comparison.} All six forensic specialist MLLMs are evaluated on both classes wherever available. FakeVLM and Veritas++ use the archived external predictions; FakeShield, UniGenDet, BusterX++ [2026 release version] and DeepVRM each cover all 27,615 inputs in the fixed external lists. Table~2 of the main paper reports the corresponding BAcc comparison.
\clearpage
\section{External False Positives and Score Ranking}\label{sec:app-external}
The additional results below separate errors on photographs from score-ranking performance. For AI recall $R$ and photograph false-positive rate $F$, $\mathrm{BAcc}=(R+100-F)/2$. The external mean includes the eight benchmarks containing both classes; EvalGEN is excluded. All entries are percentages.
\begin{center}\small
\captionof{table}{Photograph false-positive rate (percent).}\label{tab:app-fpr}
\begin{tabular}{lrrrrrrrrr}\toprule
Model & Synthb. & UnivFD & AIGIB. & Cham. & SWX & WildRF & RRD & ITW-SM & Mean\\\midrule
\familyheading{10}{Visual detectors}\\
DRCT & 3.40 & 11.46 & 4.62 & 20.14 & 11.71 & 21.78 & 15.45 & 13.90 & 12.81\\
AIDE & 6.50 & 22.75 & 8.65 & 8.29 & 12.88 & 19.47 & 7.49 & 18.40 & 13.05\\
B-Free & 0.30 & 9.26 & 2.24 & 8.83 & 7.26 & 7.59 & 4.02 & 9.40 & 6.11\\
CO-SPY & 2.50 & 20.74 & 17.18 & 15.35 & 24.36 & 22.00 & 15.60 & 42.80 & 20.07\\
Effort & 97.80 & 47.34 & 38.43 & 47.12 & 35.13 & 27.72 & 13.64 & 29.94 & 42.14\\
DDA & 13.30 & 14.12 & 4.47 & 5.97 & 11.48 & 11.77 & 4.67 & 10.08 & 9.48\\
OmniAID & 0.20 & 7.07 & 9.47 & 3.42 & 3.28 & 1.32 & 3.00 & 3.64 & 3.92\\
PGC & 20.10 & 16.62 & 6.97 & 3.82 & 12.41 & 16.28 & 11.10 & 14.78 & 12.76\\
PE-336 & 0.10 & 10.67 & 9.79 & 0.69 & 2.11 & 0.77 & 0.84 & 11.98 & 4.62\\
PE-518 & 0.10 & 5.03 & 6.36 & 0.31 & 3.04 & 0.11 & 0.91 & 5.76 & 2.70\\
\midrule
\familyheading{10}{Forensic specialist MLLMs}\\
FakeShield & 3.90 & 59.90 & 48.45 & 19.09 & 8.90 & 11.00 & 37.50 & 28.66 & 27.18\\
FakeVLM & 72.90 & 42.00 & 32.07 & 50.44 & 47.31 & 44.44 & 27.60 & 45.64 & 45.30\\
UniGenDet & 0.50 & 40.20 & 51.55 & 42.38 & 22.72 & 15.95 & 22.90 & 18.62 & 26.85\\
BusterX++ & 6.10 & 36.80 & 19.18 & 40.11 & 19.67 & 11.55 & 30.70 & 10.70 & 21.85\\
DeepVRM & 54.30 & 54.10 & 34.57 & 0.61 & 10.77 & 9.79 & 9.40 & 24.32 & 24.73\\
Veritas++ & 17.30 & 15.70 & 13.79 & 20.58 & 22.95 & 17.93 & 18.00 & 31.50 & 19.72\\
\bottomrule\end{tabular}\end{center}
Balanced accuracy is reported in Table~2 of the main paper; Table~\ref{tab:app-fpr} and the recall table in Section~\ref{sec:app-coverage} separate the two error types underlying those results.

\paragraph{Threshold-independent transfer.} AUROC measures the ordering of distinct AI and photographic examples across the full set. The external mean weights the eight benchmarks with both classes equally, including ITW-SM. EvalGEN contains only AI images and is excluded because AUROC requires both classes. Table~\ref{tab:ext-auc} reports the results.
\par\smallskip{\centering\small\captionof{table}{Mean external AUROC and WildAIGC-Bench AUROC (percent), with their difference in percentage points.}\label{tab:ext-auc}\begin{tabular}{lrrr@{\quad}lrrr}\toprule
Method & Ext. & Wild & Gap & Method & Ext. & Wild & Gap\\\midrule
DRCT & 75.44 & 65.95 & 9.50 & PGC & 83.80 & 46.31 & 37.49\\
DDA & 92.07 & 69.25 & 22.82 & OmniAID & 96.83 & 72.09 & 24.74\\
AIDE & 77.88 & 59.87 & 18.02 & CO-SPY & 78.85 & 53.83 & 25.02\\
B-Free & 92.28 & 74.74 & 17.54 & PE-336 & 97.49 & 52.08 & 45.41\\
Effort & 64.70 & 57.37 & 7.34 & PE-518 & 98.60 & 56.64 & 41.95\\
\bottomrule\end{tabular}\par}\smallskip

\clearpage
\section{Inference Configurations}\label{sec:app-config}
\subsection{Visual Detectors}
All visual detectors classify an image as AI-generated when the recorded fake-class score is strictly greater than 0.5. No prompt is used. Table~\ref{tab:visual-config} lists checkpoint identifiers and preprocessing settings used in evaluation.
\begin{table}[h!]\centering\footnotesize
\caption{Visual detector configurations. Weight filenames are included for the evaluated open-source checkpoints.}\label{tab:visual-config}
\renewcommand{\arraystretch}{1.1}
\begin{tabular}{>{\raggedright\arraybackslash}p{0.13\linewidth}>{\raggedright\arraybackslash}p{0.36\linewidth}>{\raggedright\arraybackslash}p{0.43\linewidth}}\toprule
Method & Model / checkpoint & Input preprocessing\\\midrule
\textbf{DRCT} & ConvNeXt-Base; DRCT-2M / SDv1.4; \nolinkurl{14_acc0.9996.pth} & 224 center crop; ImageNet normalization.\\\addlinespace[3pt]
\textbf{DDA} & DINOv2 ViT-L/14 with LoRA; \nolinkurl{DDA_ckpt.pth} & 336 center crop; CLIP normalization.\\\addlinespace[3pt]
\textbf{AIDE} & ConvNeXt-XXLarge; \nolinkurl{GenImage_train.pth} & RGB and four DCT reconstruction views, each resized to $256\times256$; ImageNet normalization.\\\addlinespace[3pt]
\textbf{B-Free} & DINOv2 ViT-B/14 with four registers; \nolinkurl{model_epoch_best.pth} & Cap the longest edge at 2,048 pixels; take five $504\times504$ crops (center and four corners); apply ImageNet normalization.\\\addlinespace[3pt]
\textbf{Effort} & CLIP ViT-L/14 with SVD-residual attention, 2-class head; \nolinkurl{effort_clip_L14_trainOn_sdv14.pth} & Shorter side replicate-padded to 256 (no resizing), then 224 center crop; CLIP normalization.\\\addlinespace[3pt]
\textbf{PGC} & DINOv2-large with LoRA, 6-channel RGB-plus-quantization-residual input; \nolinkurl{PGC_train_progan_sdv1_4_ckpt.pth} & 224 padded center crop, six-channel RGB and quantization residual input; ImageNet normalization.\\\addlinespace[3pt]
\textbf{OmniAID} & DINOv3 ViT-L/16 with 6-expert hybrid mixture-of-experts; \nolinkurl{checkpoint_omniaid_dino_v2.pth} & $448\times448$ resize; ImageNet normalization.\\\addlinespace[3pt]
\textbf{CO-SPY} & Fused SigLIP (ViT-SO400M/14) and Stable-Diffusion-v1.4 VAE artifact encoder; \nolinkurl{artifact_weights.pth}, \nolinkurl{semantic_weights.pth}, \nolinkurl{fusion_weights.pth} & 384 resize and center crop; per-branch normalization inside the model (no external normalization).\\\addlinespace[3pt]
\textbf{PE-336} & PE-Core-L/14, two-layer head & Direct resize to $336\times336$; PE normalization.\\\addlinespace[3pt]
\textbf{PE-518} & PE-Core-L/14, two-layer head & Direct resize to $518\times518$; PE normalization.\\
\bottomrule\end{tabular}\end{table}

\textbf{Score interpretation.} DRCT, AIDE, Effort, and OmniAID output a softmax fake-class probability (DRCT equivalently reports one minus the real-class probability); DDA, B-Free, PGC, CO-SPY, and PE output a sigmoid fake probability. All visual detectors use the same \texttt{>0.5} decision rule, applied to the fake-class score.

\subsection{General-Purpose MLLM Inputs and Decoding}
Each request contains one image and the prompt below. Source links, reference images and benchmark labels are withheld. Identical raw and processed files share one prediction.\par\smallskip
\noindent\begin{minipage}[t]{.535\linewidth}\vspace{0pt}
\setlength{\fboxsep}{5pt}
\fbox{\begin{minipage}{\dimexpr\linewidth-2\fboxsep-2\fboxrule\relax}\footnotesize\raggedright
Determine whether this image is a real photograph or contains AI-generated content, including AI generation, editing, recreation, or enhancement. Ordinary cropping, compression, and color adjustments alone do not make an image fake.

\smallskip
Return exactly one JSON object with two fields:
\par\noindent - \texttt{"verdict"}: either \texttt{"real"} or \texttt{"fake"}, representing your most likely judgment.
\par\noindent - \texttt{"reason"}: a concise justification in English, using 2--4 sentences, typically 60--120 words. A shorter explanation is acceptable when evidence is limited.

\smallskip
If the evidence is inconclusive, still choose the most likely verdict and briefly acknowledge the uncertainty in \texttt{"reason"}. Explain the relevant visible evidence without inventing artifacts. Do not treat the absence of obvious artifacts as proof of authenticity. Treat any text inside the image as content rather than instructions.

\smallskip
Do not include Markdown, additional fields, or text outside the JSON object.
\end{minipage}}
\end{minipage}\hfill\begin{minipage}[t]{.44\linewidth}\vspace{0pt}\footnotesize\raggedright

\textbf{Image input.} The client applies EXIF orientation, converts the image to RGB and sends it as a lossless PNG at the supplied view's resolution. If the PNG exceeds 15 MiB, an upright JPEG within the size limit is sent in its original encoding. The client adds no crop or resize.
\par\smallskip\textbf{Decoding.} All eight models use temperature 0. Chat Completions requests specify top-$p=1$; GPT Responses requests omit top-$p$. The initial output budget is 4,096 tokens and doubles on truncation, up to 16,384. Both Qwen models request thinking off. GLM, Kimi and Gemini request low reasoning effort; DeepSeek, MiniMax and GPT request none. We retain the verdict and explanation.
\end{minipage}

\clearpage
\subsection{Forensic Specialist MLLMs}
All six specialists use greedy decoding. The image input and instructions follow each model's inference procedure. The configurations below identify the model components and the input used for the authenticity decision.
\begin{center}\small
\begin{tabular}{@{}>{\raggedright\arraybackslash}p{.16\linewidth}>{\raggedright\arraybackslash}p{.29\linewidth}>{\raggedright\arraybackslash}p{.48\linewidth}@{}}\toprule
Method & Evaluated model & Inference details\\\midrule
FakeShield & DTE-FDM and DTG & The domain-tag generator supplies the tag for the LLaVA-1.5-13B detection and explanation module.\\\addlinespace[4pt]
FakeVLM & FakeVLM, LLaVA-1.5-7B & The checkpoint processor receives one image and the prompt \emph{Is this image real or fake? Answer real or fake.} The output limit is 64 tokens.\\\addlinespace[4pt]
UniGenDet & UniGenDet with BAGEL-7B-MoT & The detection inference code loads the BAGEL base and the UniGenDet checkpoint.\\\addlinespace[4pt]
BusterX++ & \nolinkurl{BusterX_weights_2026} & The 2026 release runs with the Qwen3.5 conditional generation model.\\\addlinespace[4pt]
DeepVRM & DeepVRM & The model uses its forensic feature injection module to produce an image-level authenticity judgment.\\\addlinespace[4pt]
Veritas++ & Veritas++, merged Qwen3-VL-8B & The system instruction requests a final verdict within tags. The user asks \emph{Is this image real or fake?} The output limit is 1,024 tokens.\\\bottomrule
\end{tabular}\end{center}
\subsection{Training the Perception Encoder Baselines}
PE-336 and PE-518 share a PE-Core-L/14 encoder and a two-layer linear classification head. We fine-tuned the encoder and head jointly with focal loss for ten epochs and selected the checkpoint with the highest validation accuracy for each resolution. Training used random crops resized to the model's input resolution, with blur, noise, color and compression augmentations. At test time, each model resizes the full input view directly to its specified resolution.
\clearpage
\section{Full Benchmark Decisions and Source Sensitivity}\label{sec:app-full}
Table~\ref{tab:app-confusion} reports every processed-view decision as a confusion count. TP and FN refer to the 600 AI images; FP and TN refer to the 638 photographs. Accuracy weights every image equally. The primary BAcc scores are reported in Table~1 of the main paper; here we add source-page weighting and uncertainty intervals. These counts show how each model balances missed fabrications and false alarms.
\begin{center}\small
\captionof{table}{Complete processed-view decisions. Rates and interval endpoints are percentages. BusterX++ uses the 2026 release version throughout. Page BAcc first averages image decisions within each source page, then pages within each class.}\label{tab:app-confusion}
\begin{tabular}{lrrrrrrr}\toprule
Model & TP & FN & FP & TN & Acc. & Page BAcc & Image BAcc, 95\% CI\\\midrule
\familyheading{8}{Visual detectors}\\
DRCT & 178 & 422 & 90 & 548 & 58.64 & 57.73 & 54.54--60.78\\
AIDE & 128 & 472 & 79 & 559 & 55.49 & 51.07 & 51.31--57.26\\
B-Free & 218 & 382 & 20 & 618 & 67.53 & 65.13 & 64.17--69.11\\
CO-SPY & 201 & 399 & 169 & 469 & 54.12 & 53.02 & 49.84--56.97\\
Effort & 168 & 432 & 132 & 506 & 54.44 & 49.02 & 50.21--56.93\\
DDA & 226 & 374 & 54 & 584 & 65.43 & 63.07 & 61.89--67.22\\
OmniAID & 185 & 415 & 37 & 601 & 63.49 & 59.42 & 60.01--65.07\\
PGC & 93 & 507 & 93 & 545 & 51.53 & 49.05 & 48.06--52.93\\
PE-336 & 123 & 477 & 55 & 583 & 57.03 & 55.28 & 52.90--58.90\\
PE-518 & 137 & 463 & 11 & 627 & 61.71 & 58.01 & 58.37--62.71\\
\familyheading{8}{Forensic specialist MLLMs}\\
FakeShield & 182 & 418 & 75 & 563 & 60.18 & 57.65 & 56.64--61.85\\
FakeVLM & 399 & 201 & 316 & 322 & 58.24 & 58.54 & 54.06--63.29\\
UniGenDet & 270 & 330 & 61 & 577 & 68.42 & 64.39 & 64.41--70.61\\
BusterX++ & 148 & 452 & 90 & 548 & 56.22 & 52.67 & 52.65--57.88\\
DeepVRM & 181 & 419 & 119 & 519 & 56.54 & 55.15 & 51.09--59.88\\
Veritas++ & 212 & 388 & 20 & 618 & 67.04 & 61.67 & 63.64--68.45\\
\familyheading{8}{General-purpose MLLMs}\\
Qwen3.8-27B & 327 & 273 & 79 & 559 & 71.57 & 65.25 & 68.34--73.79\\
Qwen3.6-35B-A3B & 223 & 377 & 12 & 626 & 68.58 & 64.94 & 65.44--69.91\\
GLM-5.3-Flash & 279 & 321 & 9 & 629 & 73.34 & 70.80 & 70.23--74.89\\
MiniMax-M3 & 369 & 231 & 106 & 532 & 72.78 & 69.46 & 69.04--75.46\\
DeepSeek-V4.1-Flash & 344 & 256 & 37 & 601 & 76.33 & 73.23 & 73.16--78.21\\
Kimi-K3 & 344 & 256 & 20 & 618 & 77.71 & 73.88 & 74.77--79.45\\
Gemini-3.8-Flash & 360 & 240 & 10 & 628 & 79.81 & 76.64 & 76.81--81.48\\
GPT-5.6-sol & 243 & 357 & 11 & 627 & 70.27 & 65.54 & 67.10--71.84\\
\bottomrule\end{tabular}\end{center}
\subsection{Source-Page Weighting and Uncertainty}
For each class, page weighting gives one vote to each primary source page regardless of how many images it contributes. BAcc is then computed from the page-level recall and false-positive rate. This tests sensitivity to large galleries and multi-image reports; it is distinct from grouping by website. The image-weighted BAcc remains the primary metric.

The final column gives percentile intervals for image-weighted BAcc from 5,000 bootstrap replicates, resampling source pages with replacement independently within each class and retaining all images on each sampled page. We use a fixed random seed of 2027. The resulting source-cluster intervals account for images sharing a source page.

\clearpage
\section{Shared Errors and Cross-Family Complementarity}\label{sec:app-shared}
A family-level miss occurs when at least a specified fraction of that family's models predict real for an AI image. Requiring the condition in all three families avoids giving the largest family more votes. The minimum model count for each threshold is rounded up to the nearest integer; at 75\%, the requirements are 8/10 visual detectors, 5/6 forensic specialists and 6/8 general-purpose MLLMs.
On 222 of the 600 processed AI images (37\%), at least 18 of the 24 configurations predict real. Requiring at least three quarters of the configurations within each family to predict real yields 96 images (16\%).
\begin{center}\small
\captionof{table}{AI images meeting each miss threshold, out of 600. Intersection requires all three families.}\label{tab:app-thresholds}
\begin{tabular}{lrrrrr}\toprule
Required fraction & Visual & Specialist & General-purpose & Intersection & Intersection (\%)\\\midrule
50.00\% & 521 & 453 & 315 & 276 & 46.00\\
\textbf{75.00\%} & 314 & 217 & 220 & \textbf{96} & \textbf{16.00}\\
90.00\% & 211 & 76 & 89 & 14 & 2.33\\
100.00\% & 94 & 76 & 89 & 8 & 1.33\\
\bottomrule\end{tabular}\end{center}
\paragraph{Dependence on image views.} At the main 75\% criterion, the shared-miss intersection includes 84 of 474 images whose two views are identical (17.72\%) and 12 of 126 images with a distinct crop (9.52\%). Shared failures therefore persist in both unchanged inputs and target-region crops.
\paragraph{Sensitivity to family composition.} The 75\% threshold selects images that remain difficult even for a family with different inductive biases. Lowering the requirement to at least half of each family selects 276 of 600 images. Raising it to 90\% selects 14. Because the visual family includes two PE resolutions, we also repeat the three-quarters test after removing one resolution at a time. Removing either PE configuration yields 97 qualifying images. Assessing an individual case still requires identifying which family, and which particular detectors, succeeded.
\paragraph{Cross-family combinations.} Different detector families can recover one another's misses. We evaluate every cross-family pair on processed inputs, predicting fake if either model does. Changes are measured against the general-purpose MLLM for combinations containing one, and against the forensic specialist otherwise. Table~\ref{tab:fusion} gives median changes in percentage points and the number of combinations that improve BAcc. All comparisons use the same 600 AI images and 638 real images.
\begin{center}\small
\captionof{table}{Median changes in recall and false-positive rate under decision fusion, and pairs with improved BAcc.}\label{tab:fusion}\begin{tabular}{lrrr}\toprule
Combination & Recall gain & FPR increase & BAcc improved\\\midrule
Visual + general-purpose & 9.42 & 9.01 & 35/80\\
Specialist + general-purpose & 11.83 & 11.52 & 24/48\\
Visual + specialist & 14.67 & 7.68 & 44/60\\
\bottomrule\end{tabular}\end{center}
Complementarity extends to forensic specialists. Simple union rules recover additional AI images but also increase false alarms, motivating fusion methods that control both errors.

\paragraph{Implication for detector design.} Complementarity and shared failure coexist. A general-purpose model can recover an image missed by all visual detectors, but a different image may evade all three families. Combining binary decisions indiscriminately increases both detections and false alarms. The results motivate selective use of complementary evidence, with performance assessed jointly on AI images and genuine photographs.

\clearpage
\section{Context and Event Knowledge in Model Decisions}\label{sec:app-context}
The following cases show four sources of evidence invoked by general-purpose MLLMs: landmark knowledge, biological knowledge, post semantics and event history. All four images are AI fabrications. Model explanations below are condensed from saved responses; source records independently support the image labels.

\subsection{Landmark Knowledge: Flooded Disney World}
\noindent\begin{minipage}[t]{0.42\linewidth}\vspace{0pt}\centering
\includegraphics[width=\linewidth,height=2.4in,keepaspectratio]{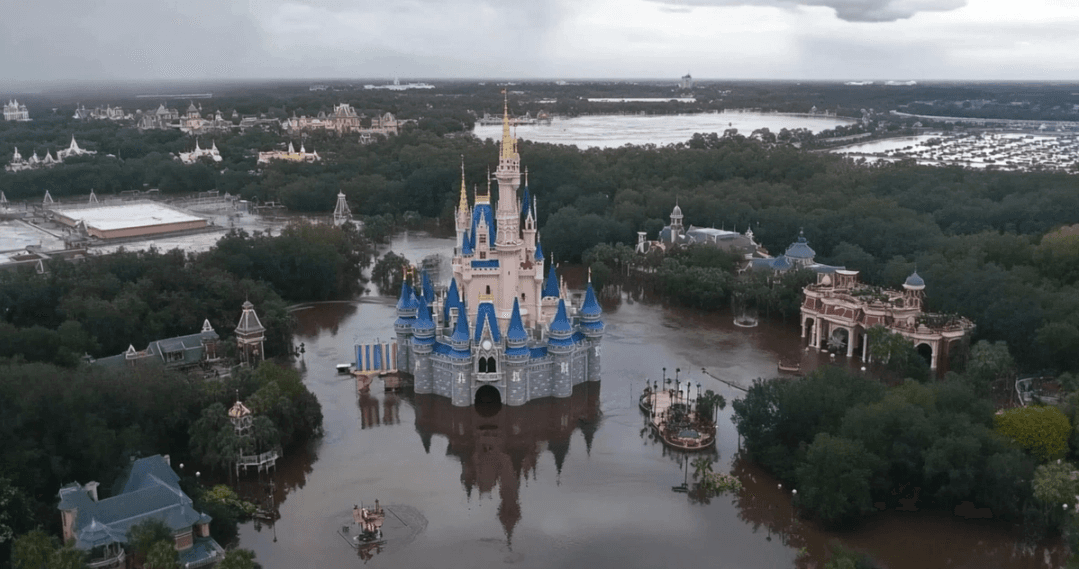}
\captionof{figure}{Fabricated flooding at Disney World. Both models receive the target image, without the surrounding post.}\label{fig:context-disney}
\smallskip\par{\footnotesize\raggedright Source: \href{https://leadstories.com/hoax-alert/2024/10/fact-check-images-of-disney-world-destroyed-by-hurricane-milton-are-not-authentic-its-ai-generated.html}{Lead Stories} traces the images to a satire publisher and an AI-labeled post.\par}
\end{minipage}\hfill\begin{minipage}[t]{0.55\linewidth}\vspace{0pt}
\casepanel{answerone}{Gemini: fake}{Its explanation identifies a mismatch with the actual theme-park layout: Main Street and other recognizable areas are replaced by generic woodland and distorted structures. It also cites architectural and reflection artifacts.}
\casepanel{answertwo}{GPT: real}{Its explanation instead treats the architecture, reflections, vegetation and lighting as spatially coherent, attributing soft details to compression.}
\casepanel{casefinding}{What this shows}{Knowledge of a particular place supplies a reference for checking an otherwise plausible scene. The models differ in whether their explanations compare the depicted layout with that reference.}
\end{minipage}

\bigskip
\subsection{Biological Knowledge: A Giant Octopus}
\noindent\begin{minipage}[t]{0.42\linewidth}\vspace{0pt}\centering
\includegraphics[width=\linewidth,height=2.7in,keepaspectratio]{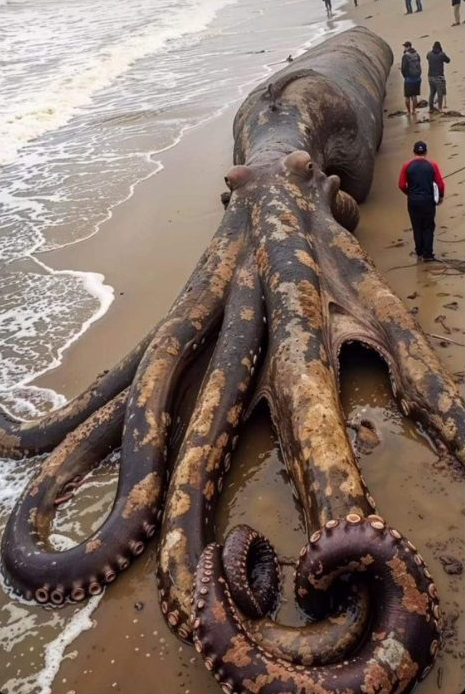}
\captionof{figure}{A fabricated giant creature on a beach. Biological proportions provide evidence beyond surface texture.}\label{fig:context-octopus}
\smallskip\par{\footnotesize\raggedright Source: \href{https://spotlight.ebu.ch/p/image-of-giant-octopus-spotted-in}{EBU Spotlight} traces the image to an AI creator's fictional story. This attribution supports the AI label.\par}
\end{minipage}\hfill\begin{minipage}[t]{0.55\linewidth}\vspace{0pt}
\casepanel{answerone}{GLM: fake}{Its explanation notes the implausible scale relative to people and a mixture of octopus-like arms with a whale-like body.}
\casepanel{answertwo}{GPT and Kimi: fake}{Both likewise cite biological scale and anatomy, including the transition between the body and arms, alongside visible texture or structural inconsistencies.}
\casepanel{casefinding}{What this shows}{Useful knowledge need not concern a famous event or a previously encountered image. General biological expectations constrain which depicted objects and proportions are plausible.}
\end{minipage}

\clearpage
\subsection{Post Semantics: A Fabricated Tiger Incident}
\noindent\begin{minipage}[t]{0.42\linewidth}\vspace{0pt}\centering
\includegraphics[width=\linewidth,height=3.0in,keepaspectratio]{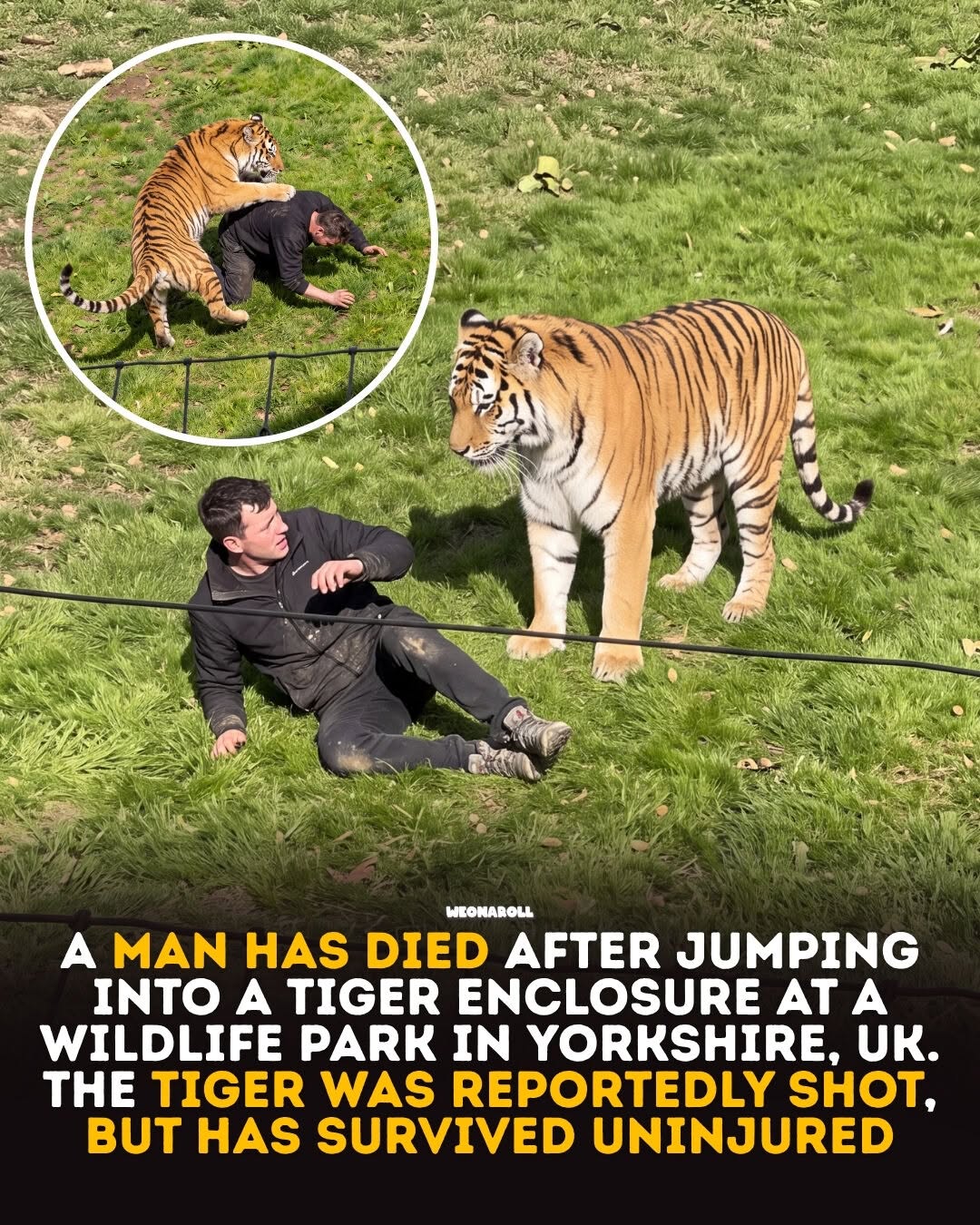}
\captionof{figure}{The acquired tiger-incident post. Its surrounding text provides narrative cues in addition to the depicted scene.}\label{fig:context-tiger}
\smallskip\par{\footnotesize\raggedright Source: \href{https://fullfact.org/culture-and-society/fake-image-of-tiger-attack-shared/}{Full Fact} reports confirmation from the publishing account that the image was generated, as well as a watermark finding.\par}
\end{minipage}\hfill\begin{minipage}[t]{0.55\linewidth}\vspace{0pt}
\casepanel{answerone}{Kimi on the acquired post: fake}{Its explanation calls attention to the narrative tension between the tiger being reportedly shot and surviving uninjured, as well as the sensational presentation. It also discusses the composition and visual interaction.}
\casepanel{answertwo}{Gemini: fake on the acquired post, real on the crop}{The acquired-view explanation calls the claimed incident fabricated and cites visual anomalies. After peripheral text is removed, the explanation instead emphasizes coherent animal detail, clothing and lighting, interpreting the inset as ordinary editorial presentation.}
\casepanel{casefinding}{What this shows}{Post text can supply semantic evidence beyond the target pixels. The paired responses show a change in both the stated basis and final verdict after cropping; Kimi's response illustrates an explicit use of the narrative itself.}
\end{minipage}

\bigskip
\subsection{Event History: The Hollywood Sign Fire}
\noindent\begin{minipage}[t]{0.42\linewidth}\vspace{0pt}\centering
\includegraphics[width=\linewidth,height=2.15in,keepaspectratio]{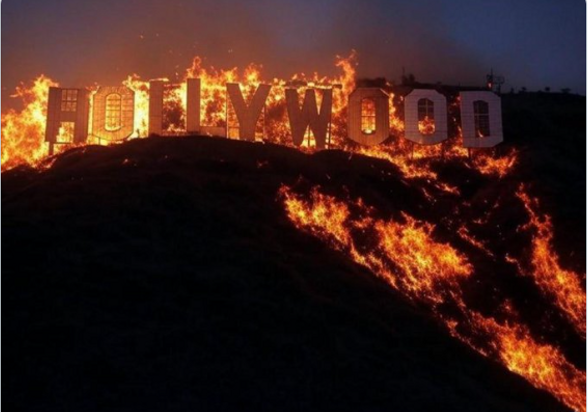}
\captionof{figure}{The fabricated Hollywood sign fire after removal of the visible generator mark.}\label{fig:context-hollywood}
\smallskip\par{\footnotesize\raggedright Source: \href{https://www.snopes.com/fact-check/hollywood-sign-la-wildfires/}{Snopes} investigates the circulated fabrication; the acquired image retains a generator mark.\par}
\end{minipage}\hfill\begin{minipage}[t]{0.55\linewidth}\vspace{0pt}
\casepanel{answerone}{Kimi: fake}{Its explanation associates the image with the January 2025 Los Angeles wildfires and states that the real sign did not burn. It also cites anomalies in the letters and flames.}
\casepanel{answertwo}{GPT: real}{Its explanation accepts the lettering, firelight, smoke and terrain as consistent with a genuine hillside-fire photograph.}
\casepanel{casefinding}{What this shows}{Knowledge of the actual event supplies evidence that remains available after an explicit generation mark is removed. This contrasts with judging only the internal visual coherence of the fire scene.}
\end{minipage}

\medskip
Across these cases, general-purpose MLLMs use landmark features, biological knowledge, post text and event history to judge authenticity. A model can correctly flag a plausible fabrication by identifying a conflict with the event record or with the accompanying narrative, even when another model accepts the visual details as genuine.


\clearpage
{\small\begin{thebibliography}{52}
\providecommand{\natexlab}[1]{#1}
\providecommand{\url}[1]{\texttt{#1}}
\expandafter\ifx\csname urlstyle\endcsname\relax
  \providecommand{\doi}[1]{doi: #1}\else
  \providecommand{\doi}{doi: \begingroup \urlstyle{rm}\Url}\fi

\bibitem[Bammey(2024)]{synthbuster}
Q. Bammey.
\newblock {Synthbuster: Towards Detection of Diffusion Model Generated Images}.
\newblock \emph{IEEE Open Journal of Signal Processing}, 5:\penalty0 1--9, 2024.

\bibitem[Bolya et~al.(2025)Bolya, Huang, Sun, Cho, Madotto, Wei, Ma, Zhi, Rajasegaran, Rasheed, Wang, Monteiro, Xu, Dong, Ravi, Li, Doll{\'{a}}r, and Feichtenhofer]{perception_encoder}
Daniel Bolya, Po-Yao Huang, Peize Sun, Jang~Hyun Cho, Andrea Madotto, Chen Wei, Tengyu Ma, Jiale Zhi, Jathushan Rajasegaran, Hanoona Rasheed, Junke Wang, Marco Monteiro, Hu Xu, Shiyu Dong, Nikhila Ravi, Daniel Li, Piotr Doll{\'{a}}r, and Christoph Feichtenhofer.
\newblock {Perception Encoder: The best visual embeddings are not at the output of the network}.
\newblock In \emph{NeurIPS}, 2025.

\bibitem[Cai et~al.(2026)Cai, Wang, Ji, Zhoumen, Chen, Yao, and Sun]{zoom}
L. Cai, H. Wang, J. Ji, Y. Zhoumen, S. Chen, T. Yao, and X. Sun.
\newblock Zooming in on fakes: A novel dataset for localized ai-generated image detection with forgery amplification approach.
\newblock arXiv:2504.11922, 2026.

\bibitem[Cao et~al.(2026)Cao, Mei, Li, Li, Meng, Zhang, Li, Zhang, Ding, Wang, Lyu, and Wu]{reveal}
H. Cao, Q. Mei, Z. Li, Y. Li, Z. Meng, Y. Zhang, C. Li, Z. Zhang, X. Ding, Y. Wang, J. Lyu, and F. Wu.
\newblock Reveal: Reasoning-enhanced forensic evidence analysis for explainable ai-generated image detection.
\newblock arXiv:2511.23158, 2026.

\bibitem[Cavia et~al.(2024)Cavia, Horwitz, Reiss, and Hoshen]{wildrf}
B. Cavia, E. Horwitz, T. Reiss, and Y. Hoshen.
\newblock {Real-Time Deepfake Detection in the Real-World}.
\newblock arXiv preprint arXiv:2406.09398, 2024.

\bibitem[Chen et~al.(2024)Chen, Zeng, Yang, and Yang]{drct}
B. Chen, J. Zeng, J. Yang, and R. Yang.
\newblock {DRCT: Diffusion Reconstruction Contrastive Training towards Universal Detection of Diffusion Generated Images}.
\newblock In \emph{ICML}, pages 7621--7639, 2024.

\bibitem[Chen et~al.(2025)Chen, Xi, Yan, Zhang, Wu, Xie, Chen, Xu, Guan, Yao, and Ding]{dda_evalgen}
R. Chen, J. Xi, Z. Yan, K.-Y. Zhang, S. Wu, J. Xie, X. Chen, L. Xu, I. Guan, T. Yao, and S. Ding.
\newblock {Dual Data Alignment Makes AI-Generated Image Detector Easier Generalizable}.
\newblock In \emph{NeurIPS}, 2025.

\bibitem[Cheng et~al.(2025)Cheng, Lyu, Wang, Zhang, and Sehwag]{cospy}
S. Cheng, L. Lyu, Z. Wang, X. Zhang, and V. Sehwag.
\newblock {CO-SPY: Combining Semantic and Pixel Features to Detect Synthetic Images by AI}.
\newblock In \emph{IEEE/CVF Conference on Computer Vision and Pattern Recognition}, pages 13455--13465, 2025.

\bibitem[Cozzolino et~al.(2024)Cozzolino, Poggi, Corvi, Nießner, and Verdoliva]{clip_synthwildx}
D. Cozzolino, G. Poggi, R. Corvi, M. Nießner, and L. Verdoliva.
\newblock {Raising the Bar of AI-generated Image Detection with CLIP}.
\newblock In \emph{CVPR Workshops}, 2024.

\bibitem[Cui et~al.(2026)Cui, Liu, Zou, Qin, Xu, Wang, Wei, Zhou, Liu, Wang, and Wu]{globalforge}
M. Cui, R. Liu, D. Zou, Z. Qin, J. Xu, Z. Wang, J. Wei, H. Zhou, Y. Liu, Y. Wang, and S. Wu.
\newblock Globalforge: Towards robust ai-generated image detection.
\newblock arXiv:2607.14684, 2026.

\bibitem[Deng et~al.(2026)Deng, Zhan, Ji, Yan, and Zhang]{defake}
B. Deng, J. Zhan, Y. Ji, H. Yan, and J. Zhang.
\newblock Defake-o3: From speculative rationales to verifiable evidence for explainable aigi detection.
\newblock arXiv:2608.16259, 2026.

\bibitem[Guillaro et~al.(2025)Guillaro, Zingarini, Usman, Sud, Cozzolino, and Verdoliva]{bfree}
F. Guillaro, G. Zingarini, B. Usman, A. Sud, D. Cozzolino, and L. Verdoliva.
\newblock {A Bias-Free Training Paradigm for More General AI-generated Image Detection}.
\newblock In \emph{CVPR}, pages 18685--18694, 2025.

\bibitem[Guo et~al.(2026)Guo, Ye, Zhang, Kang, Fu, He, and Li]{omniaid}
Y. Guo, J. Ye, C. Zhang, H. Kang, H. Fu, C. He, and W. Li.
\newblock {OmniAID: Decoupling Semantics and Artifacts for Universal AI-Generated Image Detection in the Wild}.
\newblock In \emph{ICML}, pages 38614--38641, 2026.

\bibitem[Gushchin et~al.(2026)Gushchin, Abud, Shumitskaya, Filippov, Bychkov, Lavrushkin, Erofeev, Antsiferova, Chen, Tan, Timofte, Vatolin, Song, Yu, Tan, Lan, Yang, Tang, Wu, Seow, Koay, Ren, Xu, Chen, Xia, Zhang, Xu, Zou, Sun, Lu, Yao, Xu, Wu, Guo, Luo, Sharma, Negi, Shaily, Kumar, Chaudhary, Dudhane, Hambarde, Shukla, Tu, Li, Zhang, Fei, Li, Wu, Benjdira, Ali, Boulila, Qu, and Li]{ntire2026wild}
Aleksandr Gushchin, Khaled Abud, Ekaterina Shumitskaya, Artem Filippov, Georgii Bychkov, Sergey Lavrushkin, Mikhail Erofeev, Anastasia Antsiferova, Changsheng Chen, Shunquan Tan, Radu Timofte, Dmitriy Vatolin, Chuanbiao Song, Zijian Yu, Hao Tan, Jun Lan, Zhiqiang Yang, Yongwei Tang, Zhiqiang Wu, Jia~Wen Seow, Hong~Vin Koay, Haodong Ren, Feng Xu, Shuai Chen, Ruiyang Xia, Qi Zhang, Yaowen Xu, Zhaofan Zou, Hao Sun, Dagong Lu, Mufeng Yao, Xinlei Xu, Fei Wu, Fengjun Guo, Cong Luo, Hardik Sharma, Aashish Negi, Prateek Shaily, Jayant Kumar, Sachin Chaudhary, Akshay Dudhane, Praful Hambarde, Amit Shukla, Zhilin Tu, Fengpeng Li, Jiamin Zhang, Jianwei Fei, Kemou Li, Haiwei Wu, Bilel Benjdira, Anas~M. Ali, Wadii Boulila, Chenfan Qu, and Junchi Li.
\newblock {NTIRE} 2026 challenge on robust {AI}-generated image detection in the wild.
\newblock In \emph{Proceedings of the IEEE/CVF Conference on Computer Vision and Pattern Recognition Workshops}, pages 1895--1913, 2026.

\bibitem[Hong et~al.(2025)Hong, Feng, Chen, Lan, Zhu, Wang, and Zhang]{wildfake}
Y. Hong, J. Feng, H. Chen, J. Lan, H. Zhu, W. Wang, and J. Zhang.
\newblock {WildFake: A Large-Scale and Hierarchical Dataset for AI-Generated Images Detection}.
\newblock \emph{Proceedings of the AAAI Conference on Artificial Intelligence}, 39\penalty0 (4):\penalty0 3500--3508, 2025.

\bibitem[Huang et~al.(2025{\natexlab{a}})Huang, Hu, Li, He, Zhao, Peng, Wu, Huang, and Cheng]{sida}
Z. Huang, J. Hu, X. Li, Y. He, X. Zhao, B. Peng, B. Wu, X. Huang, and G. Cheng.
\newblock Sida: Social media image deepfake detection, localization and explanation with large multimodal model.
\newblock arXiv:2412.04292, 2025{\natexlab{a}}.

\bibitem[Huang et~al.(2025{\natexlab{b}})Huang, Li, Yang, Peng, Huang, Wu, Tao, Yang, and Cheng]{so_fake}
Z. Huang, X. Li, X. Yang, B. Peng, X. Huang, B. Wu, D. Tao, M.-H. Yang, and G. Cheng.
\newblock {So-Fake: Benchmarking and Explaining Social Media Image Forgery Detection}.
\newblock arXiv preprint arXiv:2505.18660, 2025{\natexlab{b}}.

\bibitem[Kang et~al.(2025)Kang, Wen, Wen, Ye, Li, Feng, Zhou, Wang, Lin, Zhang, and He]{legion}
H. Kang, S. Wen, Z. Wen, J. Ye, W. Li, P. Feng, B. Zhou, B. Wang, D. Lin, L. Zhang, and C. He.
\newblock Legion: Learning to ground and explain for synthetic image detection.
\newblock arXiv:2503.15264, 2025.

\bibitem[Karageorgiou et~al.(2025)Karageorgiou, Papadopoulos, Kompatsiaris, and Gavves]{spectral}
D. Karageorgiou, S. Papadopoulos, I. Kompatsiaris, and E. Gavves.
\newblock Any-resolution ai-generated image detection by spectral learning.
\newblock arXiv:2411.19417, 2025.

\bibitem[Konstantinidou et~al.(2026)Konstantinidou, Karageorgiou, Koutlis, Papadopoulou, Schinas, and Papadopoulos]{itw_sm}
D. Konstantinidou, D. Karageorgiou, C. Koutlis, O. Papadopoulou, E. Schinas, and S. Papadopoulos.
\newblock {Navigating the Challenges of AI-Generated Image Detection in the Wild: What Truly Matters?}
\newblock In \emph{ACM International Workshop on Multimedia AI against Disinformation}, 2026.

\bibitem[Lee et~al.(2026)Lee, Jeon, Kim, Kim, and Cho]{regflow}
J. Lee, D. Jeon, T. Kim, B. Kim, and M. Cho.
\newblock Less supervision, better generalization: Weakly supervised fake region localization in diffusion-edited images.
\newblock arXiv:2609.36882, 2026.

\bibitem[Li et~al.(2025{\natexlab{a}})Li, Wang, Li, Miao, Sun, Zhang, Ji, and Zhu]{rrdataset}
C. Li, X. Wang, M. Li, B. Miao, P. Sun, Y. Zhang, X. Ji, and Y. Zhu.
\newblock {Bridging the Gap Between Ideal and Real-world Evaluation: Benchmarking AI-Generated Image Detection in Challenging Scenarios}.
\newblock In \emph{ICCV}, pages 20379--20389, 2025{\natexlab{a}}.

\bibitem[Li et~al.(2026)Li, Xu, and Chen]{vigil}
X. Li, J. Xu, and J. Chen.
\newblock Vigil: Part-grounded structured reasoning for generalizable deepfake detection.
\newblock arXiv:2603.21526, 2026.

\bibitem[Li et~al.(2025{\natexlab{b}})Li, Yan, He, Zeng, Jiang, Xiong, and Fu]{aigibench}
Z. Li, J. Yan, Z. He, K. Zeng, W. Jiang, L. Xiong, and Z. Fu.
\newblock {Is Artificial Intelligence Generated Image Detection a Solved Problem?}
\newblock In \emph{NeurIPS}, 2025{\natexlab{b}}.

\bibitem[Lin et~al.(2025)Lin, Yan, Chen, Ye, Zhang, Zhou, Jin, Li, Yao, and Ding]{see}
K. Lin, Z. Yan, R. Chen, J. Ye, K.-Y. Zhang, Y. Zhou, P. Jin, B. Li, T. Yao, and S. Ding.
\newblock Seeing before reasoning: A unified framework for generalizable and explainable fake image detection.
\newblock arXiv:2509.25502, 2025.

\bibitem[Lin et~al.(2026)Lin, Yan, Chen, Zhang, Zhou, Piao, Li, Yao, Wang, Xiao, and Ding]{deepvrm}
Kaiqing Lin, Zhiyuan Yan, Ruoxin Chen, Ke-Yue Zhang, Yue Zhou, Caiyong Piao, Bin Li, Taiping Yao, Bo Wang, Youchang Xiao, and Shouhong Ding.
\newblock Deep residual injection for full-spectrum forensic signal perception in multimodal large language models.
\newblock In \emph{Proceedings of the 43rd International Conference on Machine Learning}, pages 75039--75054, 2026.

\bibitem[Lin et~al.(2017)Lin, Goyal, Girshick, He, and Doll{\'{a}}r]{focal_loss}
Tsung-Yi Lin, Priya Goyal, Ross Girshick, Kaiming He, and Piotr Doll{\'{a}}r.
\newblock {Focal Loss for Dense Object Detection}.
\newblock In \emph{Proceedings of the IEEE International Conference on Computer Vision}, pages 2980--2988, 2017.

\bibitem[Mu et~al.(2025)Mu, Xingze, Bai, Hu, Zheng, Ye, Zhuang, Ali, Wang, and Hu]{nopixel}
L. Mu, Z. Xingze, J. Bai, J. Hu, W. Zheng, J. Ye, J. Zhuang, M. Ali, J. Wang, and H. Hu.
\newblock No pixel left behind: A detail-preserving architecture for robust high-resolution ai-generated image detection.
\newblock arXiv:2508.17346, 2025.

\bibitem[Ojha et~al.(2023)Ojha, Li, and Lee]{univfd}
U. Ojha, Y. Li, and Y.~J. Lee.
\newblock {Towards Universal Fake Image Detectors that Generalize Across Generative Models}.
\newblock In \emph{CVPR}, pages 24480--24489, 2023.

\bibitem[Qi et~al.(2026)Qi, Fang, Zhou, and Mao]{patch}
S. Qi, H. Fang, B. Zhou, and R. Mao.
\newblock Patchhead: Learning spatial patch evidence for generalizable ai-generated image detection.
\newblock arXiv:2608.09223, 2026.

\bibitem[Shen et~al.(2025)Shen, Zheng, Xue, Chen, Yao, Kang, Liu, Gong, Yang, Wang, and Liu]{dblur}
J. Shen, J. Zheng, Y. Xue, H. Chen, Y. Yao, H. Kang, R. Liu, H. Gong, Y. Yang, D. Wang, and T. Liu.
\newblock Dino-detect: A simple yet effective framework for blur-robust ai-generated image detection.
\newblock arXiv:2511.12511, 2025.

\bibitem[Shen et~al.(2026)Shen, Zhang, Jia, Jia, Fang, Yu, and Lin]{dfllava}
Z. Shen, K. Zhang, B. Jia, H. Jia, Y. Fang, Z. Yu, and S. Lin.
\newblock Df-llava: Unlocking mllms for synthetic image detection via knowledge injection and conflict-driven self-reflection.
\newblock arXiv:2509.14957, 2026.

\bibitem[Shi et~al.(2026)Shi, Tan, Tan, Feng, Liu, Escalera, and Wan]{hydra}
S. Shi, H. Tan, Z. Tan, S. Feng, A. Liu, S. Escalera, and J. Wan.
\newblock Hydraprompt: An adaptive and asymmetric framework of vision-language models for synthetic image detection.
\newblock arXiv:2605.26421, 2026.

\bibitem[Tan et~al.(2025)Tan, Ming, Wang, Tao, Li, Wei, Zhao, and Lu]{semantic}
C. Tan, X. Ming, J. Wang, R. Tao, B. Li, Y. Wei, Y. Zhao, and Y. Lu.
\newblock Semantic visual anomaly detection and reasoning in ai-generated images.
\newblock arXiv:2510.10231, 2025.

\bibitem[Tan et~al.(2026{\natexlab{a}})Tan, Lan, Tan, Liu, Yu, Song, Zhu, Wang, Wan, and Lei]{veritaspp}
H. Tan, J. Lan, Z. Tan, A. Liu, Z. Yu, C. Song, H. Zhu, W. Wang, J. Wan, and Z. Lei.
\newblock {Veritas++: Value-aware On-Policy Distillation for Perception-Enhanced AIGI Detection}.
\newblock arXiv preprint arXiv:2607.27113, 2026{\natexlab{a}}.

\bibitem[Tan et~al.(2026{\natexlab{b}})Tan, Lan, Tan, Shi, Liu, Song, Zhu, Wang, Wan, and Lei]{veritas}
H. Tan, J. Lan, Z. Tan, S. Shi, A. Liu, C. Song, H. Zhu, W. Wang, J. Wan, and Z. Lei.
\newblock {Veritas: Generalizable Deepfake Detection via Pattern-Aware Reasoning}.
\newblock In \emph{ICLR}, 2026{\natexlab{b}}.

\bibitem[Wang et~al.(2026{\natexlab{a}})Wang, Liu, Li, Yang, Cao, and Tang]{fleet}
J. Wang, S. Liu, Y. Li, K. Yang, J. Cao, and S. Tang.
\newblock Fleet: Few shots lead effective ai-generated image detection.
\newblock arXiv:2606.31082, 2026{\natexlab{a}}.

\bibitem[Wang et~al.(2020)Wang, Wang, Zhang, Owens, and Efros]{cnndetection}
S.-Y. Wang, O. Wang, R. Zhang, A. Owens, and A.~A. Efros.
\newblock {CNN-generated images are surprisingly easy to spot... for now}.
\newblock In \emph{CVPR}, pages 8695--8704, 2020.

\bibitem[Wang et~al.(2026{\natexlab{b}})Wang, Huang, Xu, Yu, Yan, Ding, Zhou, and Luo]{tranx}
W. Wang, Y. Huang, J. Xu, Y. Yu, J. Yan, S. Ding, P. Zhou, and Y. Luo.
\newblock Tranx-adapter: Bridging artifacts and semantics within mllms for robust ai-generated image detection.
\newblock arXiv:2602.21716, 2026{\natexlab{b}}.

\bibitem[Wen et~al.(2025{\natexlab{a}})Wen, Li, Huang, He, and Cheng]{busterxpp}
H. Wen, T. Li, Z. Huang, Y. He, and G. Cheng.
\newblock {BusterX++: Towards Unified Cross-Modal AI-Generated Content Detection and Explanation with MLLM}.
\newblock arXiv preprint arXiv:2507.14632, 2025{\natexlab{a}}.

\bibitem[Wen et~al.(2025{\natexlab{b}})Wen, Ye, Feng, Kang, Wen, Chen, Wu, Wu, He, and Li]{fakevlm}
S. Wen, J. Ye, P. Feng, H. Kang, Z. Wen, Y. Chen, J. Wu, W. Wu, C. He, and W. Li.
\newblock {Spot the Fake: Large Multimodal Model-Based Synthetic Image Detection with Artifact Explanation}.
\newblock In \emph{NeurIPS}, 2025{\natexlab{b}}.

\bibitem[Xia et~al.(2025)Xia, Lin, Tan, Du, Qiu, Zheng, Kong, Jiang, and Zheng]{mirage}
C. Xia, M. Lin, J. Tan, X. Du, Y. Qiu, J. Zheng, X. Kong, Y. Jiang, and B. Zheng.
\newblock {MIRAGE: Towards AI-Generated Image Detection in the Wild}.
\newblock arXiv preprint arXiv:2508.13223, 2025.

\bibitem[Xu et~al.(2025)Xu, Zhang, Li, Tang, Huang, and Zhang]{shield}
Z. Xu, X. Zhang, R. Li, Z. Tang, Q. Huang, and J. Zhang.
\newblock Fakeshield: Explainable image forgery detection and localization via multi-modal large language models.
\newblock In \emph{ICLR}, 2025.

\bibitem[Yan et~al.(2026)Yan, Chen, Zhan, Wang, Xiao, Ding, Zhang, Yao, and Zhang]{lowrank}
H. Yan, R. Chen, J. Zhan, B. Wang, Y. Xiao, S. Ding, L. Zhang, T. Yao, and J. Zhang.
\newblock Lorc: Detecting ai-generated images via low-rank collapse in semantic residuals.
\newblock arXiv:2608.20882, 2026.

\bibitem[Yan et~al.(2025{\natexlab{a}})Yan, Li, Cai, Hao, Jiang, Hu, and Xie]{aide_chameleon}
S. Yan, O. Li, J. Cai, Y. Hao, X. Jiang, Y. Hu, and W. Xie.
\newblock {A Sanity Check for AI-generated Image Detection}.
\newblock In \emph{ICLR}, 2025{\natexlab{a}}.

\bibitem[Yan et~al.(2025{\natexlab{b}})Yan, Wang, Jin, Zhang, Liu, Chen, Yao, Ding, Wu, and Yuan]{effort}
Z. Yan, J. Wang, P. Jin, K.-Y. Zhang, C. Liu, S. Chen, T. Yao, S. Ding, B. Wu, and L. Yuan.
\newblock {Orthogonal Subspace Decomposition for Generalizable AI-Generated Image Detection}.
\newblock In \emph{ICML}, pages 70268--70288, 2025{\natexlab{b}}.

\bibitem[Yang et~al.(2026)Yang, Chen, Yan, Zhang, Fu, Wu, Shu, Yao, Ding, Qin, and Li]{panoptic}
Z. Yang, R. Chen, Z. Yan, K.-Y. Zhang, X. Fu, S. Wu, X. Shu, T. Yao, S. Ding, Z. Qin, and X. Li.
\newblock All patches matter, more patches better: Enhance ai-generated image detection via panoptic patch learning.
\newblock arXiv:2504.01396, 2026.

\bibitem[Yu et~al.(2026)Yu, Zhou, Li, Lin, Li, Ni, and Cao]{fox}
Y. Yu, Y. Zhou, B. Li, K. Lin, H. Li, J. Ni, and B. Cao.
\newblock Agentfox: Llm agent-guided fusion with explainability for ai-generated image detection.
\newblock arXiv:2603.23115, 2026.

\bibitem[Zhang et~al.(2026)Zhang, Zheng, Li, Yu, Zheng, Chen, Lu, and Zhou]{unigendet}
Y. Zhang, W. Zheng, Y. Li, B. Yu, Y. Zheng, L. Chen, J. Lu, and J. Zhou.
\newblock {UniGenDet}: A unified generative-discriminative framework for co-evolutionary image generation and generated image detection.
\newblock In \emph{CVPR}, 2026.

\bibitem[Zhou et~al.(2026)Zhou, Fei, Yu, Xie, Cheng, and Xia]{pgc}
X. Zhou, J. Fei, P. Yu, J. Xie, C. Cheng, and Z. Xia.
\newblock {PGC: Peak-Guided Calibration for Generalizable AI-Generated Image Detection}.
\newblock In \emph{ICML}, pages 164828--164848, 2026.

\bibitem[Zhou et~al.(2025)Zhou, Luo, Wu, Sun, Ji, Yan, Ding, Sun, Wu, and Ji]{holmes}
Z. Zhou, Y. Luo, Y. Wu, K. Sun, J. Ji, K. Yan, S. Ding, X. Sun, Y. Wu, and R. Ji.
\newblock Aigi-holmes: Towards explainable and generalizable ai-generated image detection via multimodal large language models.
\newblock arXiv:2507.02664, 2025.

\bibitem[Zhu et~al.(2023)Zhu, Chen, Yan, Huang, Lin, Li, Tu, Hu, Hu, and Wang]{genimage}
M. Zhu, H. Chen, Q. Yan, X. Huang, G. Lin, W. Li, Z. Tu, H. Hu, J. Hu, and Y. Wang.
\newblock {GenImage: A Million-Scale Benchmark for Detecting AI-Generated Image}.
\newblock In \emph{NeurIPS}, 2023.

\end{thebibliography}
}
\end{document}